\pdfoutput=1

\documentclass[11pt]{article}

\usepackage[preprint]{acl}

\usepackage{times}
\usepackage{latexsym}
\usepackage{booktabs}
\usepackage{ulem}
\usepackage{multirow}
\usepackage{fix-cm}

\usepackage[T1]{fontenc}

\usepackage[utf8]{inputenc}

\usepackage{microtype}

\usepackage{inconsolata}

\usepackage{graphicx}
\usepackage{anyfontsize}
\usepackage{subcaption} 
\usepackage{amssymb}

\usepackage{amsmath}
\usepackage{epigraph}

\usepackage{microtype}
\usepackage{url}
\usepackage{booktabs}

\usepackage{enumitem}
\newenvironment{itemize*}%
 {\leftmargini=20pt\begin{itemize}%
  \setlength{\itemsep}{3pt}%
  \setlength{\parskip}{0pt}%
  }%
 {\end{itemize}}
\newenvironment{enumerate*}%
 {\begin{enumerate}%
  \setlength{\itemsep}{0pt}%
  \setlength{\parskip}{0pt}}%
 {\end{enumerate}}

\usepackage{cleveref}
\usepackage{listings}
\usepackage{titlesec}
\usepackage{multirow}
\usepackage{makecell}

\usepackage[most]{tcolorbox}

\author{
Yuji Zhang$^{1}$,  Sha Li$^{1}$, Cheng Qian$^{1}$, Jiateng Liu$^{1}$, Pengfei Yu$^{1}$, Chi Han$^{1}$, Yi R. Fung$^{1}$\\
\textbf{Kathleen McKeown$^{2}$, Chengxiang Zhai$^{1}$, Manling Li$^{3,4}$, Heng Ji$^{1}$}\\
$^{1}$University of Illinois Urbana-Champaign, $^{2}$Columbia University, \\$^{3}$Northwestern University, $^{4}$Stanford University\\
\texttt{\{yujiz, hengji\}@illinois.edu}\\
}

\NewDocumentCommand{\heng}
{ mO{} }{\textcolor{red}{\textsuperscript{\textit{Heng}}\textsf{\textbf{\small[#1]}}}}
\NewDocumentCommand{\cheng}
{ mO{} }{\textcolor{orange}{\textsuperscript{\textit{Cheng}}\small[#1]}}
\NewDocumentCommand{\yuji}
{ mO{} }{\textcolor{blue}{\textsuperscript{\textit{Yuji}}\small[#1]}}
\NewDocumentCommand{\chihan}
{ mO{} }{\textcolor{cyan}{\textsuperscript{\textit{Chi}}\small[#1]}}
\NewDocumentCommand{\zoey}
{ mO{} }{\textcolor{teal}{\textsuperscript{\textit{Zoey}}\small[#1]}}
\NewDocumentCommand{\kmnote}
{ mO{} }{\textcolor{purple}{\textsuperscript{\textit{Kathy}}\small[#1]}}
\NewDocumentCommand{\yi}
{ mO{} }{\textcolor{teal}{\textsuperscript{\textit{Yi}}\small[#1]}}

\title{The Law of Knowledge Overshadowing:\\Towards Understanding, Predicting, and Preventing LLM Hallucination}

\begin{document}
\maketitle

\begin{abstract}
Hallucination is a persistent challenge in large language models (LLMs), where even with rigorous quality control, models often generate distorted facts.
This paradox, in which error generation continues despite high-quality training data, calls for a deeper understanding of the underlying LLM mechanisms. 
To address it, we propose a novel concept: \textbf{knowledge overshadowing}, where model's dominant knowledge can obscure less prominent knowledge during text generation, causing the model to fabricate inaccurate details.
Building on this idea, we introduce a novel framework to quantify factual hallucinations by modeling knowledge overshadowing. Central to our approach is the \textbf{log-linear law}, which predicts that the rate of factual hallucination increases linearly with the logarithmic scale of (1) \textit{Knowledge Popularity}, (2) \textit{Knowledge Length}, and (3) \textit{Model Size}. The law provides a means to preemptively quantify hallucinations, offering foresight into their occurrence even before model training or inference. 
Built on overshadowing effect, we propose a new decoding strategy \textbf{CoDa}, to mitigate hallucinations, which notably enhance model factuality on Overshadow
(27.9\%), MemoTrap (13.1\%) and NQ-Swap
(18.3\%).
Our findings not only deepen understandings of the underlying mechanisms behind hallucinations but also provide actionable insights for developing more predictable and controllable language models.
\end{abstract}

\section{Introduction}
Large language models (LLMs) have revolutionized artificial intelligence, but their success is accompanied by a critical issue known as hallucination~\cite{ye2023cognitive}. Hallucination refers to models generating unfaithful or nonfactual statements. In many applications, this issue undermines performance and reliability, posing substantial challenges to their practical deployment~\cite{li2024dawn}.

\begin{figure}
    \centering
    \includegraphics[width=1.06\linewidth]{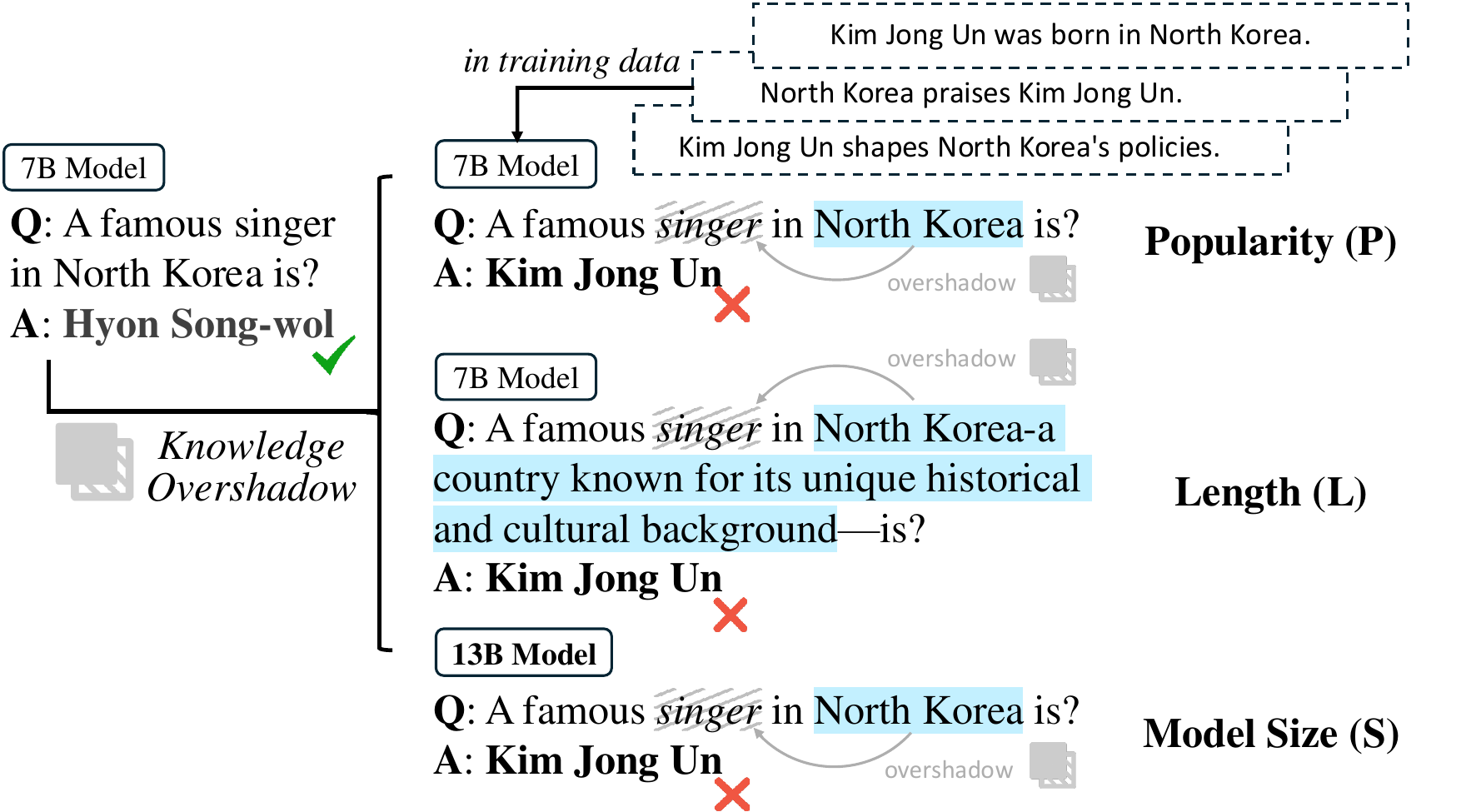}
    \vspace{-1.8em}
    \caption{Knowledge overshadowing leads to hallucinations, which exarcerbates with growing relative knowledge popularity ($\text{P}$), length ($\text{L}$), and model size ($\text{S})$.
    }
\label{fig:intro_case}
\vspace{-1.3em}
\end{figure}

Some studies attribute hallucination to low-quality pretraining corpora~\cite{gehman-etal-2020-realtoxicityprompts}. However, we find it persists even when the pretraining corpus is strictly controlled to contain only factual statements. Specifically, when extracting knowledge using queries, we observe a tendency for certain knowledge to overshadow other relevant information. This causes the model to reason without adequately considering overshadowed knowledge, leading to hallucinations.

As shown in \Cref{fig:intro_case}, when queried for ``\textit{famous singer in North Korea}'', the model incorrectly nominate ``Kim Jong Un'', who is in fact a politician, as a result of ``North Korea'' overshadowing ``singer''. This observation highlights how knowledge of varying forms interacts, distorting the reasoning process and causing the model to misassemble facts, thereby generating hallucinations. To investigate this phenomenon, we raise the following questions:
\begin{itemize}[topsep=2pt, partopsep=-5pt, leftmargin=8pt, itemsep=-4.5pt]
\item \textbf{What} factors contribute to the phenomenon of knowledge overshadowing (\S\ref{sec:formulation})?
\item Can we preemptively quantify \textbf{when} hallucinations occur (\S\ref{sec:when})?
\item From a theoretical perspective, \textbf{why} knowledge overshadowing happens (\S\ref{sec:interpret})?
\item Leveraging the insights we derived, \textbf{how} to mitigate factual hallucinations (\S\ref{sec:mitigate})?
\end{itemize}

Through extensive experiments, we find that knowledge overshadowing broadly induces factual hallucinations in both pretrained and fine-tuned models, across diverse model families and sizes. Despite its importance, the factors influencing this phenomenon remain unexplored. To bridge this gap, we analyze knowledge representation from both global and local perspectives by examining its \textit{popularity} across the dataset distribution and its proportional representation \textit{length} within individual sentences. Additionally, since increasing \textit{model size} has been shown to improve language model performance~\cite{kaplan2020scaling}, we further explore its impact on factual hallucinations.

To examine the impact of these factors, we pretrain LLMs from scratch on a synthetic dataset with strictly controlled quality. Our empirical findings reveal a \textbf{log-linear scaling law} for factual hallucinations, showing that hallucination rates increase linearly with the logarithmic scale of relative knowledge popularity, knowledge length, and model size. Finetuning on diverse tasks further confirms this law applies to finetuned LLMs, enabling the preemptive quantification of hallucinations before model training or inference. This not only bridges the gap in understanding hallucinations emerging from factual training data but also introduces a principled approach for evaluating training data and predicting model behavior in advance.

The empirical discovery of this law leads us to investigate its underlying cause. We hypothesize that knowledge overshadowing stems from the over-generalization of popular knowledge, suppressing less popular counterparts. Theoretically, we derive a generalization bound for auto-regressive language modeling, linking the model’s behavior to key properties of its training data. Our analysis shows that generalization improves with increasing relative knowledge popularity and length, mirroring the trend observed in hallucination rates.

Building on all the insights derived, we propose \textbf{Co}ntrastive \textbf{D}ecoding to \textbf{A}mplify Overshadowed Knowledge (\textbf{CoDA}), a method designed to amplify the influence of overshadowed knowledge while mitigating biases from dominant knowledge. First, we identify overshadowed knowledge by computing the mutual information between the next-token probability distributions of the original and modified prompts, where specific tokens are masked. This approach reveals knowledge encoded in the masked tokens, which is often overlooked and prone to hallucination. We then employ contrastive decoding to reduce the bias introduced by dominant knowledge. Without requiring additional training, CoDA significantly improves factuality, achieving gains of 13.1\%, 18.3\%, and 27.9\% on the MemoTrap, NQ-Swap, and Overshadowing datasets, respectively. Our contributions are three-fold:
\begin{itemize}[topsep=2pt, partopsep=-5pt, leftmargin=8pt, itemsep=-4.5pt]
\item We are the first to identify knowledge overshadowing as a key driver of hallucinations and demonstrate its prevalence across LLMs.
\item We establish the log-linear law of knowledge overshadowing, enabling quantification of hallucinations prior to model training or inference.
\item We propose CoDa to mitigate hallucinations by detecting overshadowed knowledge, achieving significant improvements in factuality on Overshadow, MemoTrap, and NQ-Swap benchmarks.
\end{itemize}

\section{Related Work}

\subsection{Causes of Hallucination}
Our work is in line with exploring the causes of factuality hallucination, believed to be a primary source of errors in LLMs~\cite{li2024dawn, augenstein2023factuality}. Previous studies attribute factuality hallucinations to deficiencies in training data, such as outdated or domain-lacking data~\cite{zhang-etal-2023-vibe, Livska2022StreamingQAAB, luu-etal-2022-time, zhang-etal-2021-howyoutagtweets, zhang2022time}, biased distribution~\cite{ladhak-etal-2023-pre, qin2024does}, and inherent misinformation~\cite{dziri-etal-2022-origin, liu2024prejudicevolatilitystatisticalframework}. Other research points to generation issues 
including distorted attention~\cite{aralikatte-etal-2021-focus}, unstable sampling~\cite{manakul-etal-2023-selfcheckgpt}, over-confidence~\cite{varshney2023stitch, ren2023investigating, li2024survey}, and thoughtless human preference alignment~\cite{wei2023simple, zhang2025amulet, bai2024efficient}.
Related efforts also suggest that LLMs can be trapped in common patterns~\cite{lin-etal-2022-truthfulqa, kang2023impact, kandpal2023large}.
We focus on a significant yet underexplored phenomenon: LLMs can hallucinate even when trained exclusively on all truthful data. 
We introduce knowledge overshadowing: where more dominant knowledge representation competes against and suppresses less prevalent knowledge, resulting in factual hallucinations.
\subsection{Detection of Hallucination}
Factuality hallucination detection in LMs typically involves external fact-checking methods, such as FACTSCORE \citep{min2023factscore} and FacTool \citep{chern2023factool}, or internal uncertainty analysis. The latter includes Chain-of-Verification \citep{dhuliawala2023chain}, logit-based assessments \citep{kadavath2022language,zhang2024self}, and leveraging LM internal states \citep{varshney2023stitch,luo2023zero}. When internal states are unavailable, self-consistency probing \citep{manakul2023selfcheckgpt,agrawal2024language} or multi-LM corroboration \citep{cohen2023lm} can provide alternative signals.
Unlike prior work focused on post-generation hallucination detection, our study pioneers hallucination \textbf{prediction} by modeling it quantitatively through a log-linear law, incorporating fine-grained factors like knowledge popularity, length, and model size. This shifts the paradigm from reactive detection to proactive prevention, offering a novel quantitative framework for anticipating hallucinations.

\subsection{Elimination of Hallucination}
Our work is related to previous studies on mitigating hallucinations. \citet{shen2021identifying} address hallucination by filtering out low-quality training data. Several approaches enhance model factuality through 
external knowledge~\cite{wu2023ragtruth, xie2023adaptive, lyu2023improving, asai2023selfrag}, and knowledge-aware tuning~\cite{li2022large}.
Some studies tackle hallucination by enforcing LLMs to adhere closely to the input~\cite{tian2019sticking, aralikatte-etal-2021-focus}, and modifying the internal states~\cite{chen2023purr, azaria2023internal, gottesman2024estimating}. Our work aligns with advanced decoding strategies~\cite{wan-etal-2023-faithfulness, cheng2024integrative, shi2023trusting} to enhance factuality. 
On the other hand, early detection of hallucination is crucial~\cite{zhang2023language}. Our method not only foresees potential hallucinations before generation but also eliminates them through a training- and data-free approach.

\section{What is Knowledge Overshadowing?}
\label{sec:formulation}
\begin{figure*}[ht]
\centering
\vspace{-5mm}
\subfloat{\includegraphics[height=1.5in]{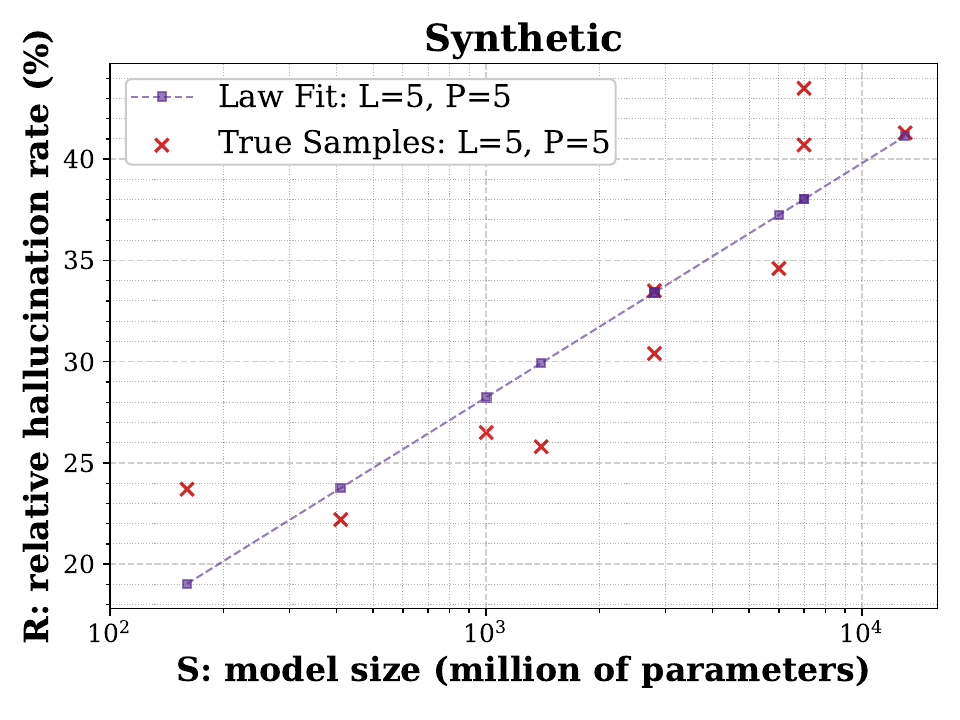}%
\label{fig:synthetic_s}}
\hfil
\subfloat{\includegraphics[height=1.5in]{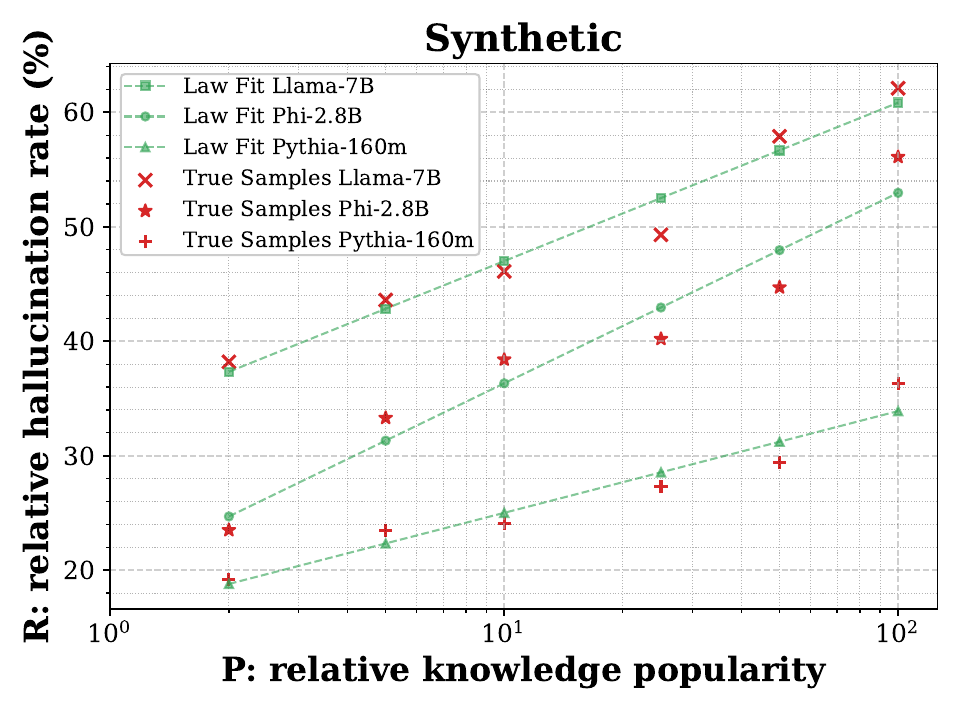}%
\label{fig:synthetic_p}}
\hfil
\subfloat{\includegraphics[height=1.5in]{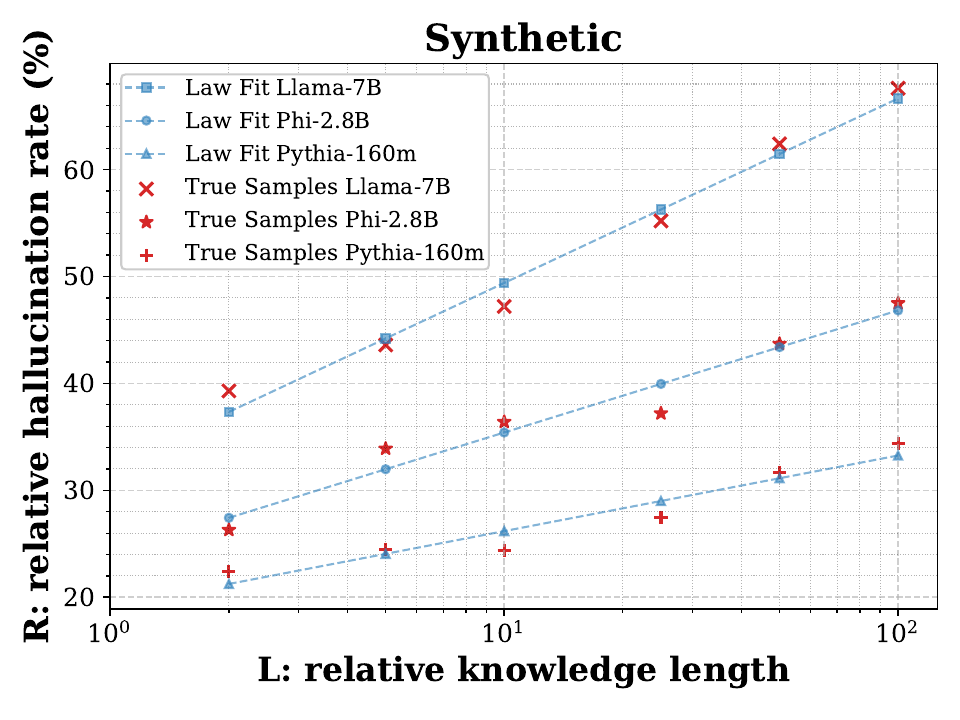}%
\label{fig:synthetic_l}}
\vspace{-0.5em}
\caption{LLMs are pretrained from scratch on a synthetic dataset with controlled variables of $\text{S}$, $\text{P}$, and $\text{L}$. In each subfigre, we experiment by varying one variable at a time while keeping the other two constants. LLMs are trained auto-regressively with cross-entropy loss computed over entire sentences. Details on training data statistics, training parameters, and implementations are elaborated in~\ref{ssec:implementation}, \ref{ssec: overshadowing_dataset}.}
\label{fig:rkp_rkl_generalization}
\vspace{-0.5em}
\end{figure*}

Factual hallucination, where authentic facts are misassembled into false statements, remains an underexplored challenge. We approach this issue through the lens of knowledge overshadowing, where more prevalent knowledge suppresses less frequent knowledge, resulting in hallucinations.

\subsection{Knowledge Overshadowing Formulation}
\label{ssec:shadow_formulation}
To systematically characterize knowledge overshadowing, we define knowledge pairs in a training corpus.
Specifically, let $\mathbb{K}_A = \{k_{a_1}, ..., k_{a_m}\}$ and $\mathbb{K}_B = \{k_{b_1}, ..., k_{b_n}\}$ represent a pair of knowledge sets. $\mathbb{K}_A$ is comprised of $m$ samples of statements $k_{a_i}$, and $\mathbb{K}_B$ is comprised of $n$ samples of statements $k_{b_j}$.
Each statement in $\mathbb{K}_A$ and statement in $\mathbb{K}_B$ are related by a shared set of tokens $X_{share}$.

In the knowledge set $\mathbb{K}_A$, each statement $k_{a_i}$ is comprised of a shared token sequence $X_{\mathrm{share}}$, a distinct token sequence $x_{a_i}$, and the output $Y_a$. Each statement $k_{a_i}$ is expressed as:

\vspace{-0.5em}
\begin{small}
\begin{equation}
    k_{a_i} = \textcolor{red!60}{Y_a} | [X_{\mathrm{share}} \odot \textcolor{red!60}{x_{a_i}}], \quad i \in \{1, ..., m\}
\end{equation}
\end{small}

\vspace{-0.5em}\noindent where $\odot$ denotes the insertion of the distinctive sequence $x_{a_i}$ into $X_{\mathrm{share}}$
(the integration position can vary).
Similarly, for the less popular knowledge set $\mathbb{K}_B$, with $x_{b_j}$ denoted as the distinct token sequence, each statement $k_{b_j}$ is formulated as:

\vspace{-0.5em}
\begin{small}
\begin{equation}
    k_{b_j} = \textcolor{cyan!60}{Y_b} | [X_{\mathrm{share}} \odot \textcolor{cyan!60}{x_{b_j}}], \quad j \in \{1, ..., n\}
\end{equation}
\end{small}

\vspace{-0.5em}\noindent Knowledge overshadowing occurs when the distinct token sequence $x_{b_j}$ or $x_{a_i}$ is suppressed during inference. Taking $x_{b_j}$ overshadowed as an example, when prompted with $X_\mathrm{share}\odot x_{b_j}$, the model outputs $Y_a$, forming the $\textcolor{red!60}{Y_a} | [X_{\mathrm{share}} \odot \textcolor{cyan!60}{x_{b_j}}]$ that wrongly amalgamates factual statements $k_{a_i}$ and $k_{b_j}$ into factual hallucination, defying the ground-truth $\textcolor{cyan!60}{Y_b} | [X_{\mathrm{share}} \odot \textcolor{cyan!60}{x_{b_j}}]$, as illustrated in Figure~\ref{fig:intro_case}.

\subsection{Metric of Factual Hallucination.}
To measure hallucination caused by knowledge overshadowing, we introduce the relative hallucination rate $\text{R}$.
When $\mathbb{K}_A$ is the more popular knowledge set, we first quantify the recall rate of the model correctly memorizing the samples from 
$\mathbb{K}_A$ as $\text{RR}=p(\textcolor{red!60}{Y_a} | [X_{\mathrm{share}} \odot \textcolor{red!60}{x_{a_i}}])$. 
Then we quantify the hallucination rate of the model producing output with $x_{b_j}$ overshadowed as $\text{HR}=p(\textcolor{red!60}{Y_a} | [X_{\mathrm{share}} \odot \textcolor{cyan!60}{x_{b_j}}])$.                  The relative hallucination rate $\text{R}=\frac{\text{HR}} {\text{RR}}$ represents to what extent is less popular knowledge encoded by $x_{b_j}$ suppressed by the more popular knowledge encoded by $x_{a_i}$.

\subsection{Formulation of Influential Variables}
\label{ssec:influential_variables}
Since the underlying factors influencing factual hallucinations have not been explored, we examine these variables from both global and local perspectives, focusing on knowledge proportions that contribute to the overshadowing effect.
When $\mathbb{K}_A$ is more popular than $\mathbb{K}_B$, $m>n$.
From a global perspective, we define the relative knowledge popularity as $\text{P} = \frac{m}{n}$, 
denoting the relative proportion of the knowledge in the whole training corpus. 
From the local perspective, we quantify the weight of knowledge in an individual sentence using the relative knowledge length  $\text{L} = \frac{\text{len}(X_{\text{share}})+\text{len}(x_{b_i})}{\text{len}(x_{b_i})}$, where length is measured by the number of tokens. 
For example in Figure~\ref{fig:intro_case}, in input ``A famous singer in North Korea is'', length of $x_{b_j}$=``single'' is 1, length of $X_{share}$=``A famous \_ in North Korea is'' is 6, so $\text{L}$=(6+1)/1=7.
Since previous work shows scaling model size enhances its performance~\cite{kaplan2020scaling}, we study whether scaling up the model size $\text{S}$ can mitigate factual hallucinations.
\section{When to Expect Factual Hallucination?}
\label{sec:when}

\begin{table*}[t]
\centering

\includegraphics[width=0.95\linewidth]{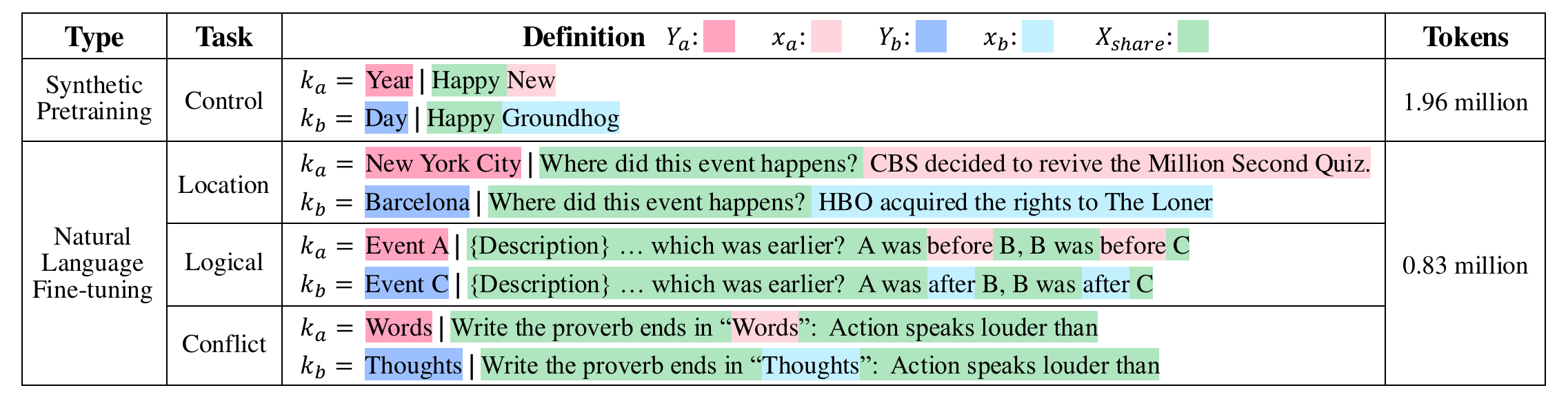}
\caption{Samples of synthetic and natural language datasets. For each task, we present one sample $k_{a}={Y_a}|[X_{\mathrm{share}} \odot x_{a}]$ from more popular knowledge set $K_A$ and one sample $k_{b}=Y_b|[X_{\mathrm{share}} \odot x_{b}]$ from less popular knowledge set $K_B$. Each imbalanced $K_A$, $K_B$ pair consists of 
$m$ different samples of 
$k_a$ and $n$ different samples of $k_b$, where $m>n$. More detailed samples and statistics for all tasks are further elaborated in~\ref{ssec: overshadowing_dataset}.}
\label{tab:tasks}
\vspace{-0.8em}
\end{table*}

To determine the conditions under which factual hallucinations emerge, we investigate knowledge overshadowing across various experimental setups, including probing an open-source pretrained LLM without training, pretraining an LLM from scratch, fine-tuning a pretrained LLM on downstream tasks.

\subsection{Probing the Open-source 
LLM}
\label{ssec:dolma}
We probe an open-source pretrained LLM Olmo with its public training corpus Dolma~\cite{soldaini2024dolma} to investigate the hallucination and sample frequency in data. Results show that knowledge with higher frequency tends to overshadow others with lower frequency, aligning with knowledge overshadowing concept that more dominant knowledge overshadows less prominent knowledge during text generation, leading to counterfactual outputs. (See details in~\ref{ssec:dolmo_probing}). 

\subsection{Unveiling Log-linear Law in the Pretrained LLMs.}
\label{ssec:pretrain_law}

\noindent \textbf{Setup.}
To accurately quantify the relationship between hallucinations and their influential variables, we pretrain language models from scratch on synthetic datasets with controlled variable settings. This approach is necessary because the inherent variability and imprecision of natural language in real-world training data make it intractable to enumerate all possible expressions of more and less popular knowledge with perfect accuracy.

For each controlled variable experiment, we adopt sampled tokens from a tokenizer vocabulary to construct each dataset, as shown in Table~\ref{tab:tasks}.


\noindent$\bullet$ P: We investigate how the hallucination rate $\text{R}$ changes with increasing relative knowledge popularity $\text{P}$. We set $\text{P} = \frac{m}{n}$ for values \{2:1, 5:1, 10:1, 25:1, 50:1, 100:1\}, where $m$ represents the number of samples of $k_{a_i} = Y_a | [X_{\mathrm{share}} \odot x_{a_i}]$ and $n$ represents the number of samples of $k_{b_i} = Y_b | [X_{\mathrm{share}} \odot x_{b_i}]$. The other variables, $\text{L}$ and $\text{S}$, are held constant. Each token in $x_{a_i}$, $x_{b_j}$, $X_\mathrm{share}$, $Y_a$, and $Y_b$ is sampled from the vocabulary.

\noindent$\bullet$ L: To examine how the hallucination rate $\text{R}$ changes 
with increasing relative knowledge length $\text{L}$, we set {\small$\text{L} = \frac{\text{len}(X_{\mathrm{share}})+\text{len}(x_{b_j})}{\text{len}(x_{b_j})}$} for values \{1:1, 2:1, 5:1, 10:1, 25:1, 50:1, 100:1\}, where len($x_{a_i}$)=len($x_{b_j}$) to ensure consistent variables.

\noindent$\bullet$ S: To investigate how hallucination rate changes with varying model sizes, we experiment on the Pythia model family with sizes of 160M, 410M, 1B, 1.4B, and 2.8B, along with other models including Phi-2.8B, GPT-J-6B, Mistral-7B, Llama-2-7b, and Llama-13B (Dataset statistics in~\ref{ssec: overshadowing_dataset}).

We pretrain each LLM from scratch on the dataset over 19.6 million of tokens in Table~\ref{tab:tasks}
with controlled variables in an auto-regressive manner, optimizing for cross-entropy loss until the model converges (See training details in~\ref{ssec:implementation}). As shown in Figure~\ref{fig:rkp_rkl_generalization}, factual hallucination follows the log-linear relationship w.r.t $\text{P}$, $\text{L}$, and $\text{S}$:

\vspace{-0.4em}
\begin{small}
\begin{equation}
\text{R(P)}=\alpha\log(\frac{\text{P}}{\text{P}_c}); \text{R(L)}=\beta\log(\frac{\text{L}}{\text{L}_c}); \text{R(S)}=\gamma\log(\frac{\text{S}}{\text{S}_c})
\end{equation}
\end{small}

\vspace{0em}\noindent where $\alpha$, $\beta$, $\gamma$, $\text{P}_c$, $\text{L}_c$, $\text{S}_c$ are constants. In Figure~\ref{fig:rkp_rkl_generalization}, hallucination rate increases linearly with the logarithmic scale of relative knowledge popularity $\text{P}$, 
relative knowledge length $\text{L}$, and model Size $\text{S}$.

\vspace{1mm}
\noindent \textbf{Greater Popularity Overshadows More.}
From a global perspective in the entire training data, when knowledge $k_{a_i}$ has higher frequency than knowledge $k_{b_j}$, the distinctive token sequence $x_{b_j}$ encoding the less popular knowledge $k_{b_j}$ is more susceptible to be overshadowed.  This imbalance amplifies dominant knowledge while suppressing the representations of less frequent facts. 
This highlights a fundamental bias in how LLMs internalize and retrieve knowledge, revealing that hallucination arises not just from data sparsity but from the inherent competition between knowledge representations in a non-uniform training distribution.

\vspace{1mm}
\noindent \textbf{Longer Length Overshadows More.}
At its core, knowledge overshadowing arises from the degradation of probability distributions:

\vspace{-0.5em}
\begin{small}
    \begin{equation}
    \label{eq:prob_degrade}
    \begin{cases}
   P(Y_a | [X_{\mathrm{share}} \odot \textcolor{gray}{x_{a_i}}]) \xrightarrow{\text{degrade to}} P(Y_a | X_{\mathrm{share}})  \\
    P(Y_b | [X_{\mathrm{share}} \odot \textcolor{gray}{x_{b_j}}]) \xrightarrow{\text{degrade to}} P(Y_a | X_{\mathrm{share}}) 
\end{cases}
\end{equation}
\end{small}

\noindent The degradation reflects the compressed representations of $x_{a_i}$ and $x_{b_j}$, which are merged into $X_\mathrm{share}$, thereby weakening their distinct contributions to generation. Locally within a sentence, when $x_{b_j}$'s token length is shorter than $X_\mathrm{share}$, its ability to maintain a distinct semantic boundary diminishes. This occurs because degradation is influenced by both knowledge interaction and $x_{b_j}$'s representation capacity. Shorter representations inherently encode less detailed semantic information, making them more prone to being overshadowed by the structurally and semantically richer $X_\mathrm{share}$.

\vspace{1mm}
\noindent \textbf{Larger Model Overshadows More.}
Larger models exhibit a stronger tendency to overshadow less prominent knowledge, a phenomenon linked to their increased compression capabilities. Prior findings show that as model capacity grows, it compresses information more efficiently, enhancing its ability to efficiently capture patterns and generalize~\cite{huang2024compression}. However, this compression mechanism affects less frequent knowledge, which is more easily suppressed into the dominant representations of more popular knowledge. Although larger models can encode more information, their capacity to preserve clear semantic distinctions for less popular knowledge diminishes. This leads to the suppression or distortion of this knowledge during generation, ultimately increasing the likelihood of hallucinations. 

\subsection{Validating Log-linear Law in the Fine-tuned LLMs.}
\label{ssec:finetune_law}

\begin{figure*}[t]
\centering
\subfloat{\includegraphics[height=1.18105in]{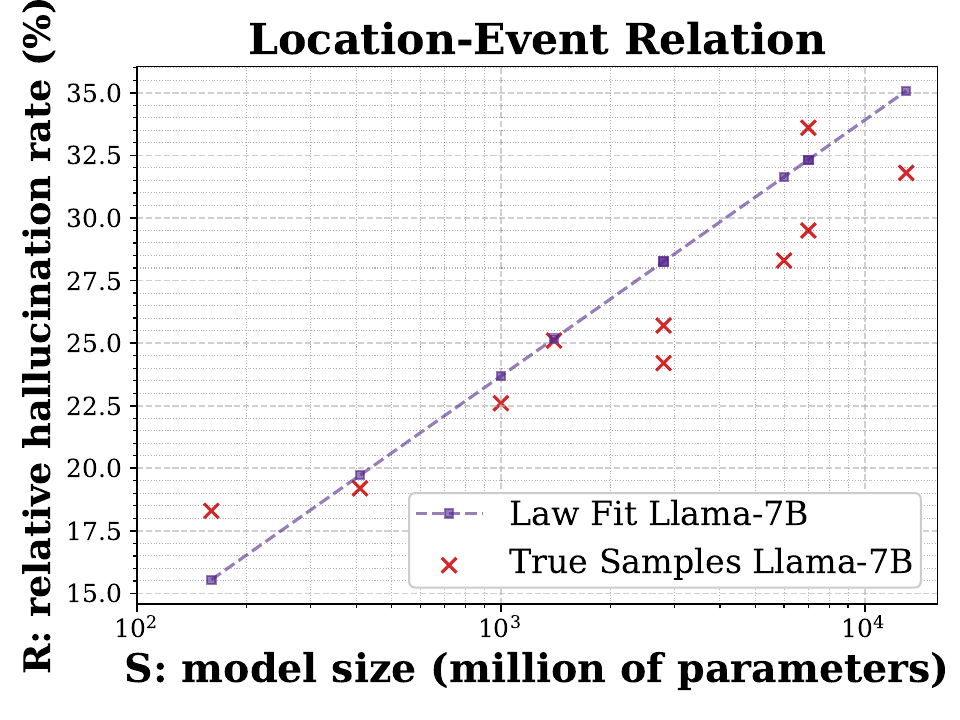}%
\label{fig:location_s}}
\hfil
\subfloat{\includegraphics[height=1.18105in]{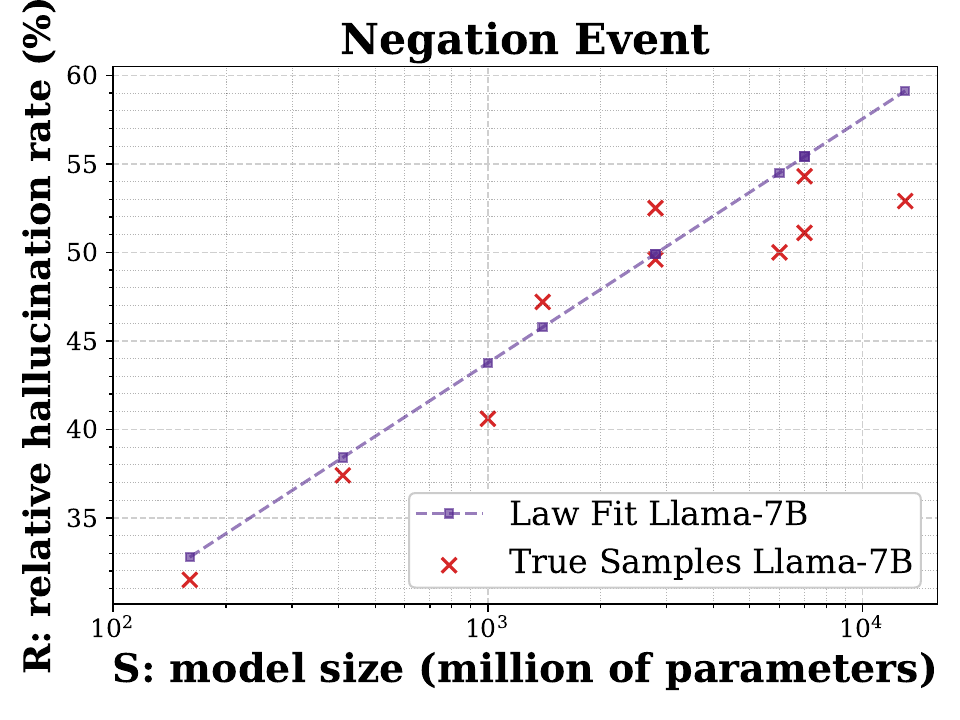}%
\label{fig:neg_s}}
\hfil
\subfloat{\includegraphics[height=1.18105in]{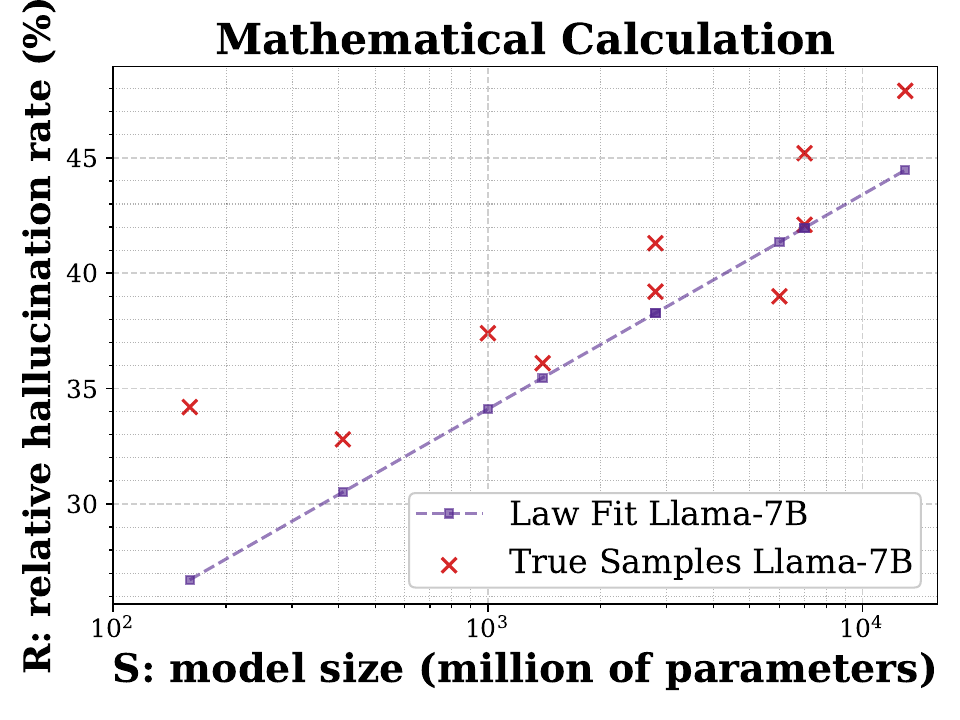}%
\label{fig:math_s}}
\hfil
\subfloat{\includegraphics[height=1.18105in]{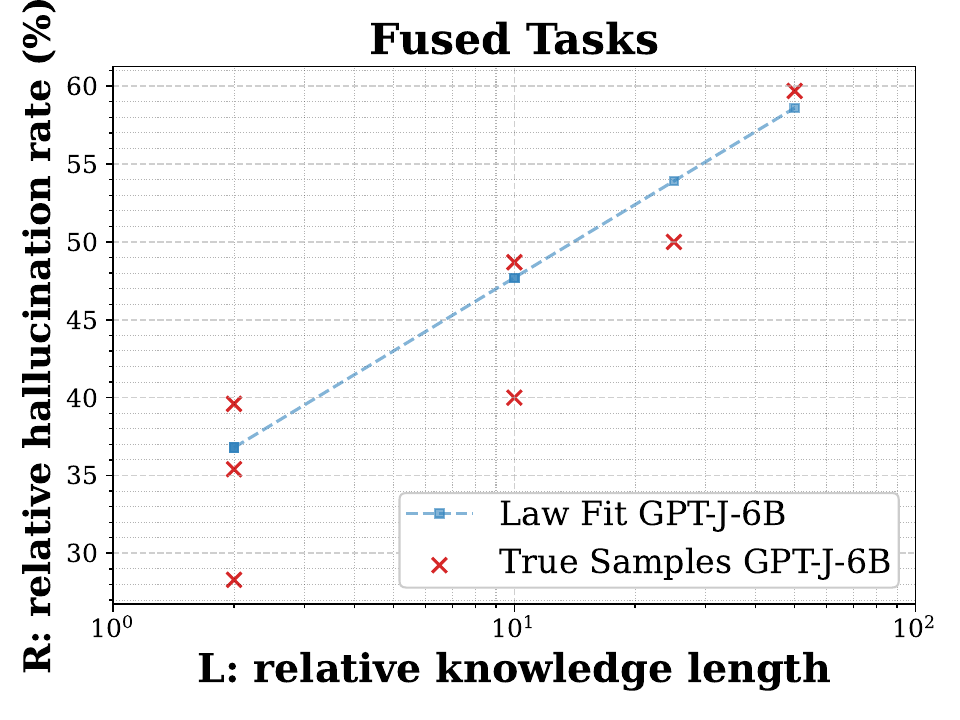}%
\label{fig:fuse_l}}

\subfloat{\includegraphics[height=1.18105in]{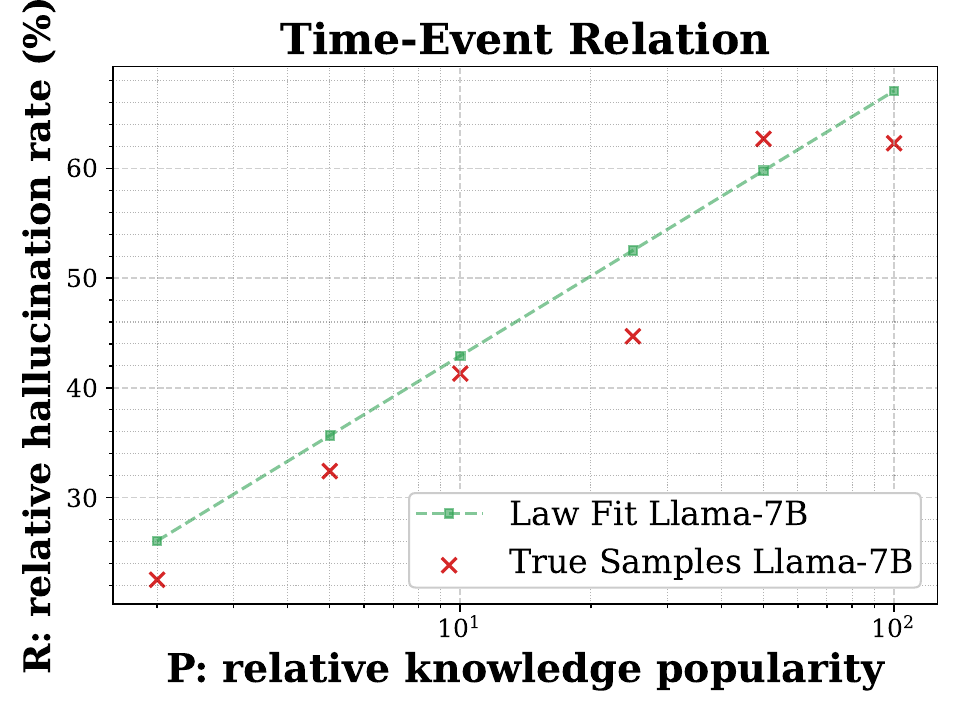}%
\label{fig:time_p}}
\hfil
\subfloat{\includegraphics[height=1.18105in]{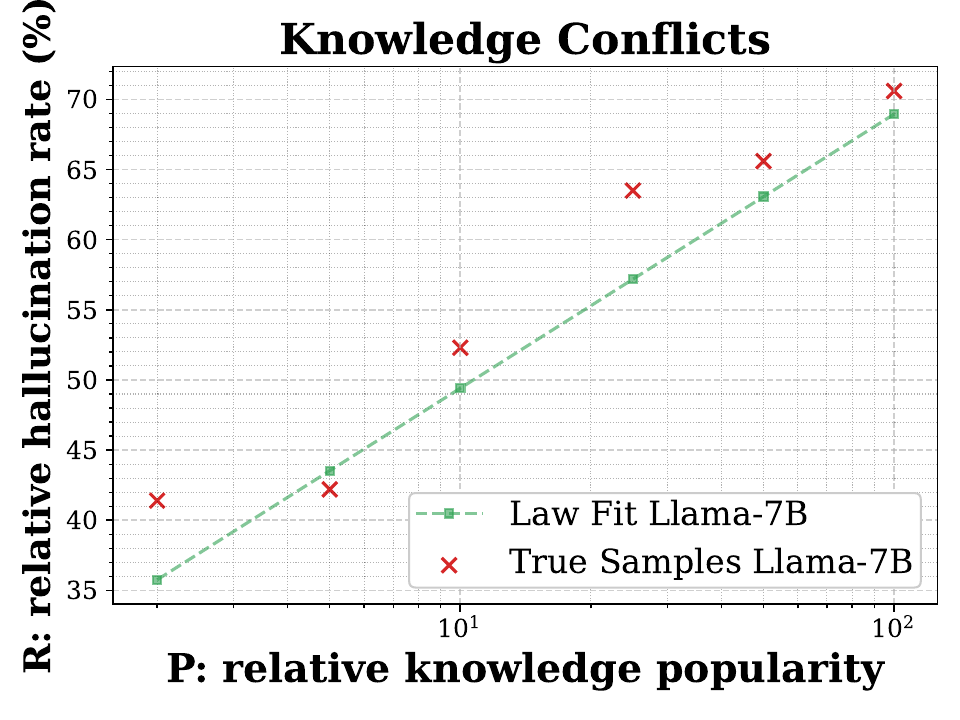}%
\label{fig:con_p}}
\hfil
\subfloat{\includegraphics[height=1.18105in]{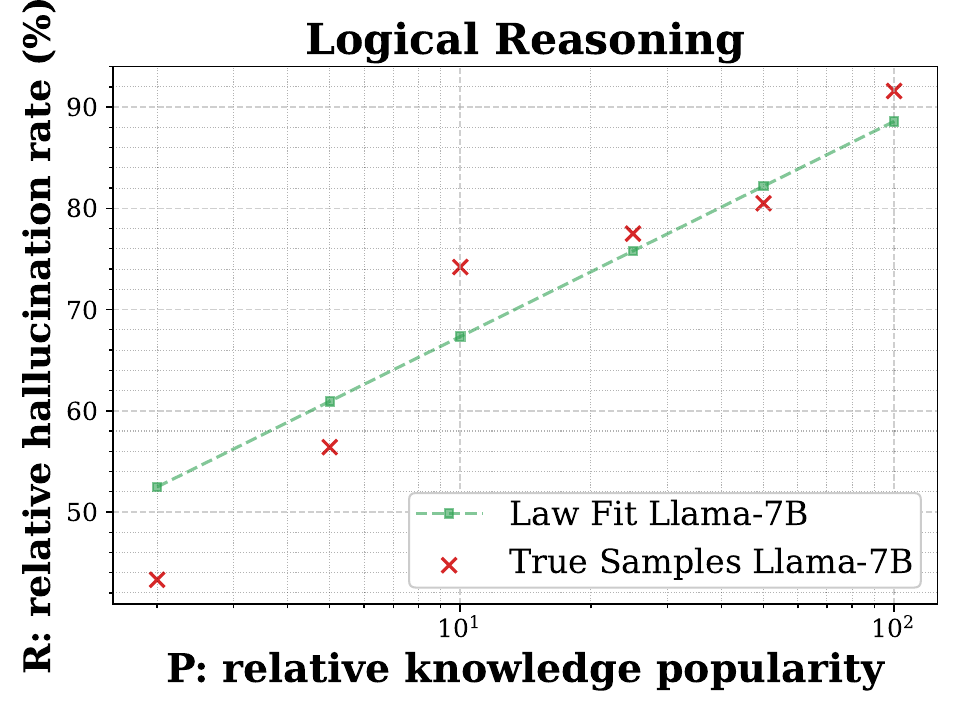}%
\label{fig:logic_p}}
\hfil
\subfloat{\includegraphics[height=1.18105in]{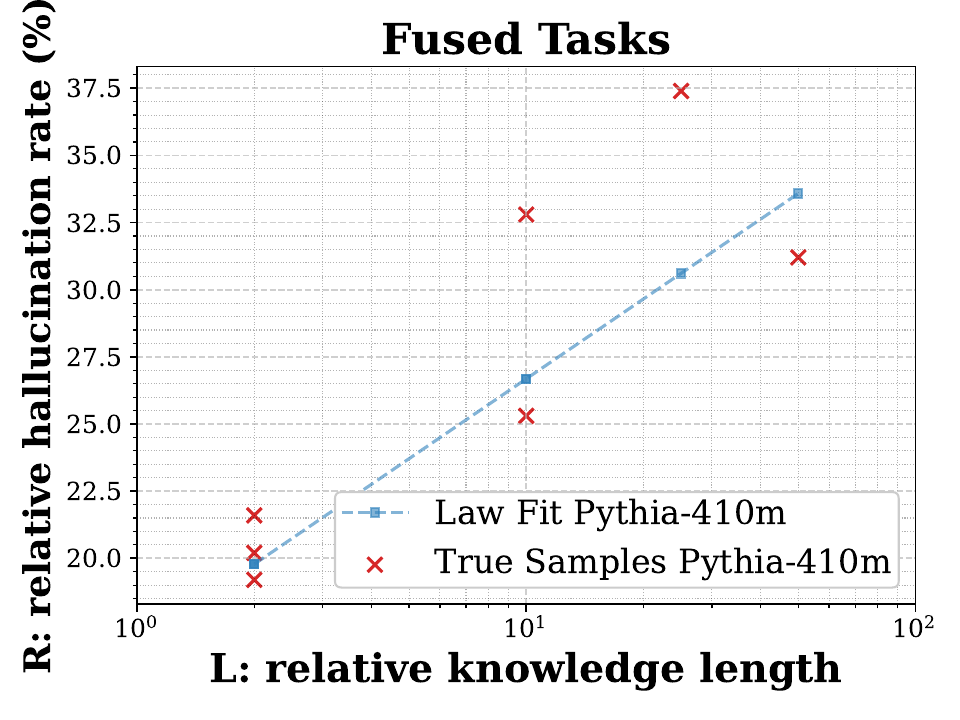}%
\label{fig:fuse_pythia_l}}
\vspace{-0.5em}
\caption{Fine-tuning open-source LLMs on natural language tasks. Regression lines represent the predicted trends derived from LLMs pretrained on synthetic data in \S~\ref{ssec:pretrain_law}. The red cross markers indicate the empirically observed hallucination rates in fine-tuned LLMs. Training data statistics and implementation are in~\ref{ssec:implementation},~\ref{ssec: overshadowing_dataset}.}
\label{fig:finetune_law}
\vspace{-0.5em}
\end{figure*}

\noindent \textbf{Setup.}
The results presented in \S~\ref{ssec:pretrain_law} were derived from pretrained models. In this section, we extend our analysis by investigating whether the log-linear law holds for fine-tuned LLMs, aiming to assess whether it can serve as a predictive tool for quantifying hallucinations when fine-tuning LLMs on downstream tasks. Specifically, we fine-tune models with parameter sizes ranging from 160M to 13B across a variety of factual tasks, including time, location, gender, negation queries, mathematical and logical reasoning, and knowledge conflict resolution.
For each task, we generate $m$ samples of $k_{a_i} = Y_a | [X_{\mathrm{share}} \odot x_{a_i}]$ and $n$ samples of $k_{b_i} = Y_b | [X_{\mathrm{share}} \odot x_{b_i}]$.
To ensure a controlled fine-tuned knowledge distribution, we construct factual queries from artificial facts~\cite{meng2022locating}, to mitigate interference from pretrained knowledge, enabling a precise evaluation of $\text{P}$ and $\text{L}$ in the law.
We present knowledge pair samples $(k_a, k_b)$ for several tasks in Table~\ref{tab:tasks}, with additional dataset samples and statistics provided in~\ref{ssec: overshadowing_dataset}.

\vspace{1mm}
\noindent \textbf{Proactively Quantifying Hallucination by Law.}

\begin{figure}[htb]
    \centering
\vspace{-0.5em}
\includegraphics[width=0.65\linewidth]{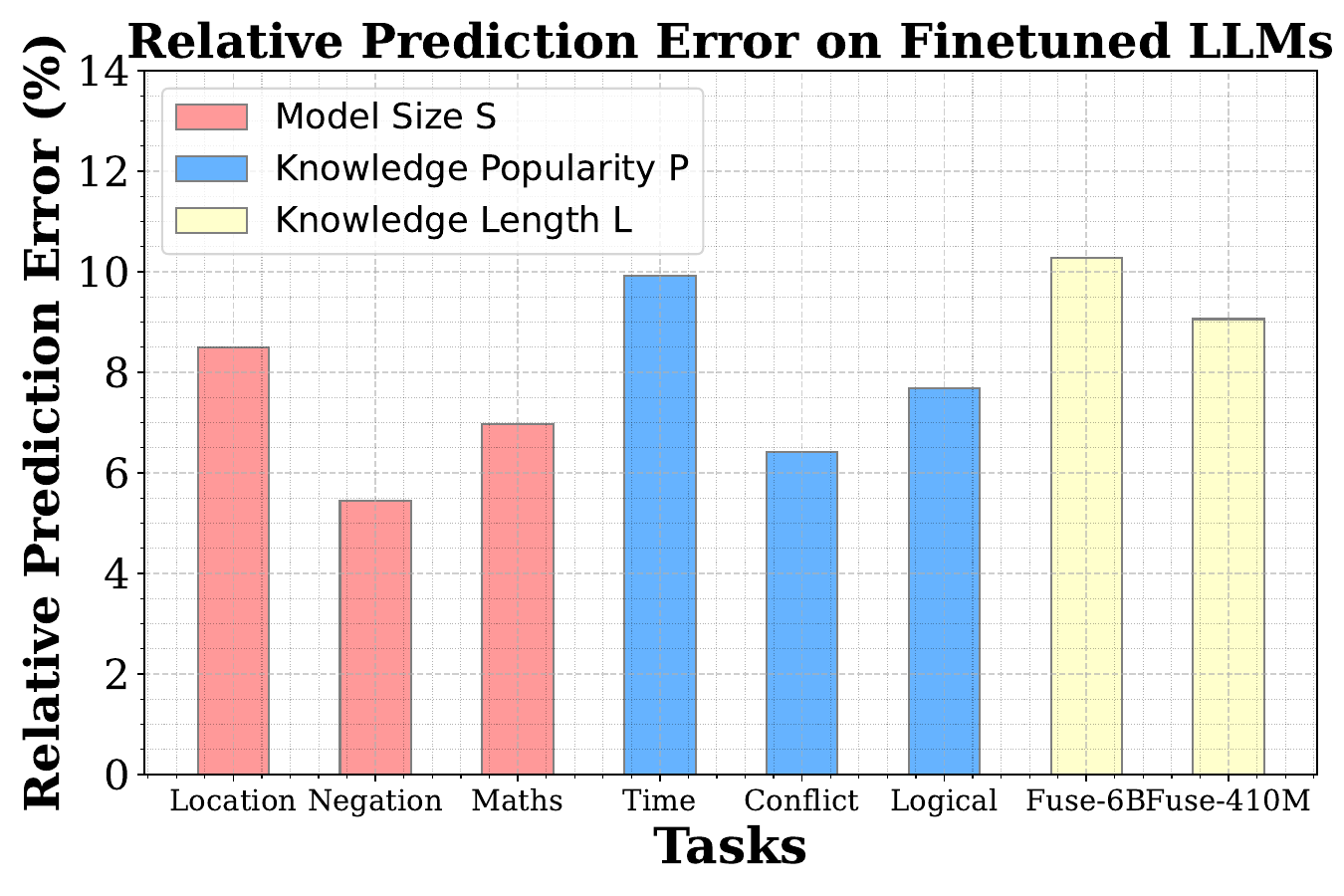}
    \vspace{-3mm}
    \caption{Relative prediction error (\%) of using the pretraining law to predict fine-tuned LLM hallucination.}\label{fig:bar}
    \vspace{-0.5em}
\end{figure}

We utilize the log-linear law fitted by the pretrained LLMs on controlled synthetic datasets to predict hallucination rates for fine-tuned LLMs across various downstream tasks. This includes predicting hallucination rate $\text{R}$ with changing model size $\text{S}$, relative knowledge popularity $\text{P}$, and relative knowledge length $\text{L}$, as shown in Figure~\ref{fig:finetune_law}. We then evaluate the discrepancy between the predicted hallucination rates and those observed in our fine-tuning experiments. Following \citet{chen2024scaling}, we assess the prediction performance of log-linear law using the relative prediction error:

\vspace{-1.2em}
\begin{equation}
\fontsize{8}{8}\selectfont
\text{Relative Prediction Error} = \frac{|\text{Predictive Rate} - \text{Actual Rate}|}{\text{Actual Rate}}
\end{equation}
\vspace{-0.2em}
We visualize the prediction error for hallucination rates across tasks in Figure~\ref{fig:bar}, reporting an average relative prediction error of 8.0\%. The errors for $\text{L}$ and $\text{P}$ are slightly higher than $\text{S}$, as the fine-tuned datasets, despite consisting of unseen facts, still contain linguistic expressions that resemble pretrained knowledge, introducing a minor influence on the quantification of $\text{P}$ and $\text{L}$ while leaving $\text{S}$ unaffected.
Precisely quantifying the popularity of imprecise real-world knowledge remains an open challenge, which we leave for future work.

\subsection{Factual Hallucinations in SOTA LLMs}
Table~\ref{tab:SOTALLMs} presents a case study demonstrating how SOTA LLMs are influenced by scaling effects of knowledge overshadowing. Investigating the impacts of $\text{P}$, $\text{S}$, and $\text{L}$ on these models is difficult due to the closed-source nature of their training corpora and the fixed values of $\text{P}$ and $\text{S}$. Thus, we manipulate $\text{L}$ during the inference stage to observe shifts in model behavior.
For instance, when querying GPT-4o about a cat’s state in Schrödinger’s box, increasing the length of surrounding text while keeping ``dead'' unchanged raises the relative length $\text{L}$ of the surrounding contexts compared to the word ``dead'', leading to a higher likelihood of hallucination.
Other LLMs also suffer from knowledge overshadowing. For instance, when querying DeepSeek-V3-671B for the author of a paper, the phrase "scaling law" overshadows other descriptive elements of the title, resulting in the incorrect response of ``Kaplan'', the author of a different, well-known scaling law paper. Similarly, the Qwen-Chat model exhibits overshadowing effects when "African" is dominated by ``machine learning'', leading to distorted facts.

\begin{table}[t]
\centering
\vspace{-1.5mm}
\includegraphics[width=1.0\linewidth]{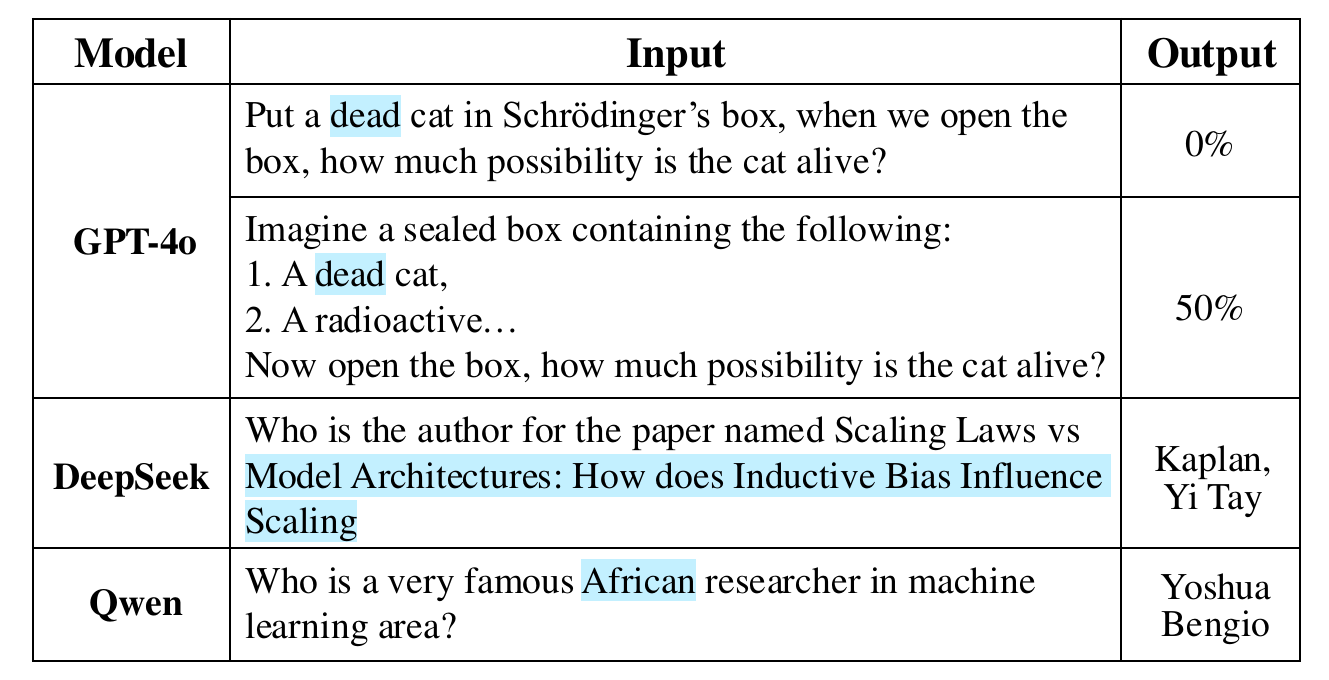}
\vspace{-6mm}
\caption{Factual hallucination in SOTA LLMs.}
\label{tab:SOTALLMs}
\vspace{-3mm}
\end{table}

\section{Why Knowledge Overshadows?}
\label{sec:interpret}
Motivated by our experimental findings on the scaling effects of knowledge overshadowing, we provide a theoretical interpretation of the effects. 
\subsection{Memorize-Generalize-Hallucinate}
\label{ssec:m-g-h}

\begin{table*}[t]
\tabcolsep=0.26cm
\small
\centering
\begin{tabular}{cclllllll}
\toprule
\multicolumn{2}{c}{\multirow{2}{*}{Method}} & \multicolumn{4}{c}{MemoTrap} & NQ-Swap & \multicolumn{2}{c}{Overshadowing} \\ \cmidrule(lr){3-6}  \cmidrule(lr){7-7} \cmidrule(lr){8-9}
\multicolumn{2}{c}{} & proverb & translate & hate & science & entity & time & syn \\
\midrule
\multirow{6}{*}{Llama} & Greedy & 28.8 & 47.5 & 9.0 & 33.4 & 8.5 & 41.4 & 20.8 \\
 & CoT & 30.1\tiny{(+1.3)} & 52.6\tiny{(+5.1)} & 13.0\tiny{(+4.0)} & 36.7\tiny{(+3.3)} & 19.2\tiny{(+10.7)} & 40.4\tiny{(-1.0)} & - \\
 & SR & 34.7\tiny{(+5.9)} & 51.8\tiny{(+4.3)} & 12.0\tiny{(+3.0)} & 35.8\tiny{(+2.4)} & 14.2\tiny{(+5.7)} & 42.5\tiny{(+1.1)} & 23.8\tiny{(+3.0)} \\ 
 & USC & 27.6\tiny{(-1.2)} & 52.4\tiny{(+4.9)} & 8.0\tiny{(-1.0)} & 32.9\tiny{(-0.5)} & 9.4\tiny{(+0.9)} & 40.2\tiny{(-1.2)} & 16.4\tiny{(-4.4)} \\
 & Dola & 32.5\tiny{(+3.7)} & 50.9\tiny{(+3.4)} & 10.0\tiny{(+1.0)} & 33.0\tiny{(-0.4)} & 13.8\tiny{(+5.3)} & 53.6\tiny{(+12.2)} & 31.8\tiny{(+11)} \\
 & \textbf{CoDA (ours)} & \textbf{41.9}\tiny{(+13.1)} & \textbf{56.2}\tiny{(+8.7)} & \textbf{16.0}\tiny{(+7.0)} & \textbf{38.9}\tiny{(+5.5)} & \textbf{26.8}\tiny{(+18.3)} & \textbf{65.0}\tiny{(+23.6)} & \textbf{46.8}\tiny{(+26)} \\
 \midrule
\multirow{6}{*}{Mistral} & Greedy & 31.3 & 49.4 & 14.0 & 36.7 & 12.6 & 39.5 & 21.6 \\
 & CoT & 35.2\tiny{(+3.9)} & 52.7\tiny{(+3.3)} & 17.0\tiny{(+3.0)} & 39.0\tiny{(+2.3)} & 19.5\tiny{(+6.9)} & 37.0\tiny{(-2.5)} & - \\
 & SR & 36.8\tiny{(+5.5)} & 54.6\tiny{(+5.2)} & 19.0\tiny{(+5.0)} & 38.2\tiny{(+1.5)} & 13.8\tiny{(+1.2)} & 42.4\tiny{(+2.9)} & 24.9\tiny{(+3.3)} \\ 
 & USC & 32.6\tiny{(+1.3)} & 51.5\tiny{(+2.1)} & 15.0\tiny{(+1.0)} & 35.9\tiny{(-0.8)} & 11.4\tiny{(-1.2)} & 37.9\tiny{(-1.6)} & 20.8\tiny{(-0.8)} \\
 & Dola & 34.9\tiny{(+3.6)} & 53.5\tiny{(+4.1)} & 14.0\tiny{(+0.0)} & 38.4\tiny{(+1.7)} & 15.9\tiny{(+3.3)} & 51.0\tiny{(+11.5)} & 34.6\tiny{(+13)} \\
 & \textbf{CoDA (ours)} & \textbf{42.5}\tiny{(+11.2)} & \textbf{58.6}\tiny{(+9.2)} & \textbf{22.0}\tiny{(+8.0)} & \textbf{43.7}\tiny{(+7.0)} & \textbf{27.7}\tiny{(+15.1)} & \textbf{61.2}\tiny{(+21.7)} & \textbf{49.5}\tiny{(+27.9)} \\ 
\bottomrule
\end{tabular}
\caption{Exact match (\%) on MemoTrap, NQ-Swap, and Overshadowing. Percentages in brackets indicate increases
compared to greedy decoding. Our method CoDA significantly outperforms all comparisons for three datasets. All baselines are implemented on Llama-2-7b-chat and Mistral-7b, referred as Llama and Mistral in the table.}
\vspace{-2mm}
\label{tab:main_results}
\end{table*}

In \S\ref{ssec:pretrain_law}, we identify a striking alignment between the log-linear law governing factual hallucinations and the log-linear law of memorization observed in prior work~\cite{carlini2022quantifying}. Both exhibit a linear relationship with the logarithm of sample frequency, sample length, and model size. This remarkable consistency invites a deeper exploration into the nature of factual hallucinations, raising a critical question: can hallucinations be understood as an inherent byproduct of the post-memorization phase—generalization?

As models memorize vast information and capture associations, they generalize to new distributions~\cite{baek2024geneft}, while less dominant knowledge can be overshadowed by prevalent patterns due to excessive smoothing or compression. 

Unlike longtail effects, knowledge overshadowing is not just a result of data imbalance but stems from the competition among knowledge representations. Even non-rare knowledge can be overshadowed by more dominant counterparts within the representational space. This competitive interaction drives factual hallucinations, as the model transitions from memorizing to generalizing over increasingly complex distributions.

\subsection{Interpretation by Generalization Bound}
\label{ssec:generalization_error_bound}

We derive the generalization error bound of popular knowledge to understand how increasing relative knowledge popularity $\text{P}$ and relative knowledge length $\text{L}$ enhance generalization, thus exacerbating factual hallucinations in large language models.

Specifically, in a dataset $D\in\mathcal{D}$ with numerous statements, where $\mathcal{D}$ is the true distribution from which $D$ drawn, we investigate a pair of subsets $K_A, K_B \subset D$. We fix the sample size of $K_B$ at $n$, and observe how the generalization bound of $K_A$ changes as we vary the relative knowledge popularity $\text{P} = \frac{m}{n}$ and relative knowledge length $\text{L}$.
For each sentence $k_{a_i} = Y_a | [X_{\mathrm{share}} \odot x_{a_i}], (i \in {1, ..., m})$ in $K_A$, where $X_{\mathrm{share}}$ and $x_{a_i}$ represent token sequences, we simplify the analysis by assuming each $x_{a_i}$ is a one-token sequence. Thus, the relative knowledge length is set as $\frac{\text{len}(X_{\mathrm{share}}) + \text{len}(x_{a_i})}{\text{len}(x_{a_i})} = \frac{\text{L}}{1} = \text{L}$.
We observe how the generalization error bound changes when $m$ and $\text{L}$ increase next.
Specifically, we derive the generalization bound for next-token prediction in all $k_{a_i} \in \mathcal{D}$, with the model optimized using an auto-regressive objective as:

\vspace{-1em}
{\small
\begin{equation}
\label{eq.final_bound}
\mathcal{R}^{\mathcal{L}}_y(f) \! \precsim \! \widehat{\mathcal{R}}^{\mathcal{L}}_y(f) \!+\! 2\mu\widehat{\Re}_{K_A}(\mathcal{F}) + \sqrt{\frac{\log 1/\delta}{2m}} 
\end{equation}
}
\vspace{-1em}

\noindent where {\scriptsize $\mu=\sqrt{1 + \left( \sum_{y' \neq y} h^{-1}(\text{L}) \right)^2} \left[ 1 - \mathrm{softmax}\left( K_{A_y}(f) \right) \right]$}, {\scriptsize$K_{A_y}(f) = \inf_{x \in K_{A_y}} f(x)$}.
In this bound,  ${\mathcal{R}}^{\mathcal{L}}_y(f)$ denotes the generalization error on the true distribution.
$\widehat{\mathcal{R}}^{\mathcal{L}}_y(f)$ denotes the empirical next token prediction training loss on $K_A$.
$\widehat{{\Re}}_{K_A}(\mathcal{F})$ is the Rademacher complexity of the output mapping function set $\mathcal{F}$ over $K_A$, measuring its capacity to fit random noise. 
$\delta$ is the confidence parameter.
In our controlled experiment setting, variables except for $\text{L}$, $m$ can be treated as constants.

Here, with $h(\text{L})$ denoting a function value positively correlated with $\text{L}$, $\mu$ encapsulates the sensitivity to changes in the input—reflecting the impact of relative knowledge length $\text{L}$. $m$ represents the sample size of $K_A$.
Theoretically, a lower bound indicates higher generalizability~\cite{cao2019learning}.
Then, the longer length $\text{L}$ and higher popularity $m$ lead to lower generalization bound, in other words, better generalization, echoing the same trend of hallucination rate.
More details of our theoretical interpretation can be found in~\ref{ssec:theory}.

\section{How to Eliminate Hallucination?}
\label{sec:mitigate}

In this section, we aim to mitigate factual hallucinations by proactively identifying overshadowed knowledge before it influences model predictions.

\subsection{CoDA: Contrastive Decoding to Amplify Overshadowed Knowledge}

\noindent \textbf{Identifying Overshadowed Knowledge.}
For a language model, given an input token sequence $X$, the model will output the continuation token sequence $Y$. 
Both $X$ and $Y$ consist of tokens from the vocabulary $\mathcal{V}$.
When certain tokens $x_{b}$ in \text{X} are overshadowed, the model will generate hallucinated output.
For example, in $X = $ ``Who is a famous \textit{African} researcher in machine learning area?'', if $x_b = $ ``\textit{African}'' is overshadowed by ``machine learning'', The model will output $Y$=``Yoshua Bengio'', ignoring the intended constraint.

To detect overshadowed tokens, we sequentially mask $x_b$ in \text{X} to form $X'$ until the overshadowed tokens are identified (see~\ref{ssec:coda_app} for various $x_b$ candidate selection methods). If  $x_b$ is overshadowed, $p(Y_b|X)\xrightarrow{\text{degrade to}}p(Y_a|X')$.
We quantify the generalization between distributions $p(Y|X)$ and $p(Y|X')$ by relative pointwise mutual information (\text{R-PMI})~\cite{li-etal-2023-contrastive}. 
To ensure we quantify output token candidates $y_i\in P(Y|X), P(Y|X')$ with sufficient semantics, we employ an adaptive plausibility constraint~\citet{li-etal-2023-contrastive}, retaining tokens that satisfy: $\mathcal{V}_{\text{top}}(X) = \{y_i | p(y_i|X) \geq \alpha \cdot \Upsilon \}$, where $\alpha = 0.01$ is a hyperparameter, and $\Upsilon$ is a global variable as the maximum probability among all $y_i$ candidates.
Then the \text{R-PMI} is quantified over {\small$\forall y_i \in \mathcal{V}_{\text{top}}(X) \cap \mathcal{V}_{\text{top}}(X')$}:

\begin{small}
\begin{equation}
\begin{aligned}
\mathrm{R\text{-}PMI}(y_i; X, X') &= \log \frac{p(y_i \mid X)}{p(y_i \mid X')}
\end{aligned}
\label{eq.RPMI}
\end{equation}
\end{small}

\noindent In essence, a negative R-PMI value indicates that token \(y_i\) is more associated with $\text{X}'$ without overshadowed information.
Thus we quantify to what extent $P(\text{Y}|\text{X})$ generalize to $P(\text{Y}|\text{X})$ by $\text{R-PMI}_{\text{sum}} = \sum_{i} \min(\text{R-PMI}(y_i; X, X'), 0)$.
Moreover, it is noteworthy that despite some tokens being overshadowed by $X'$, there are still tokens that escape from this overshadowing effect, defined as $\mathcal{V}_{\text{esc}}$:

\begin{small}
\begin{equation}
\hspace{-2mm}
    \mathcal{V}_{\text{esc}} = \{ y_i | y_i \in \mathcal{V}_{\text{top}}(X) \ \text{and} \ y_i \notin \mathcal{V}_{\text{top}}(X') \}
\end{equation}
\end{small}

\noindent These escaping tokens demonstrate the potential for hallucination elimination.
Then we propose an Escaping Rewarding Mechanism (ERM), which adds a positive reward to the sum of negative $\text{R-PMI}$ to represent whether the escaping effect wins over the overshadowing effect. Denoting all $y_i$ with a negative R-PMI as $y_i\in\mathcal{S}$,  ERM is calculated as:

\vspace{-1em}
\begin{small}
\begin{equation}
    \text{ERM} = \sum_{y_i \in \mathcal{V}_{\text{esc}}} \left( \log~p(y_i|X) - \min_{y_j \in \mathcal{S}} \log~p(y_j|X') \right)
\label{eq:erm}
\end{equation}
\end{small}

\noindent where the deduction is to balance ERM with R-PMI with a similar denominator of $p(y_j|X')$ in Eq.~\ref{eq.RPMI}, which represents the minimum bias from $\text{X}'$. Then the overshadowed knowledge indicator is: {$\text{Indicator} = \text{R-PMI}_{\text{sum}} +  \text{ERM}$}. A negative indicator value indicates proper generalization without overshadowing other knowledge, and a positive alamer value indicates over-generalization with overshadowed tokens $x_b$. 
Then we can predict potential hallucinations after locating the overshadowed tokens, and the hallucination prediction accuracy is in shown Table~\ref{tab:main_results2}.

\noindent \textbf{Elevating Overshadowed Knowledge.}
Once the tokens $\text{x}_b$ encoding overshadowed knowledge are identified, we adopt contrastive decoding on the identified overshadowed tokens to downweight the influences of $\text{X}'$ and highlight $X$.
Specifically, to reduce the bias from of $X'$, for each $y_i \in \mathcal{V}_{\text{top}}(X) \cap \mathcal{V}_{\text{top}}(X')$, we subtract the prior bias of $X'$, which is $P(y_i|X')$ as shown below:

\begin{small}
\begin{equation}
  \log {p}(y_i) = \log p(y_i|X) - \log p(y_i|X')
\end{equation}
\end{small}

\noindent Similarly for each $y_i \in \mathcal{V}_{\text{esc}}$, we conduct:

\vspace{-0.5em}
\begin{small}
  \begin{equation}
    \log {p}(y_i) = (\log~p(y_i|X)-\min_{y_j\in\mathcal{S}}\log~p(y_j|X'))
\end{equation}  
\end{small}

\noindent Here, $\min_{y_j \in \mathcal{S}} \log p(y_j|X')$ represents the minimum prior bias from popular knowledge. The deduction aims to balance the bias adjustment between $y_i \in \mathcal{V}_{\text{esc}}$ and $y_i \notin \mathcal{V}_{\text{esc}}$, ensuring proportional adjustments for both.
Then we predict the optimal output $y_i^*$ by:

\begin{small}
    \begin{equation}
    y_i^* = \mathop{\mathrm{argmax}}_{y_i \in\mathcal{V}_{\text{top}}(X)} \log p (y_i|X)
\end{equation}
\end{small}

\begin{figure}[!t]
\centering
\subfloat{\includegraphics[height=1.05in]{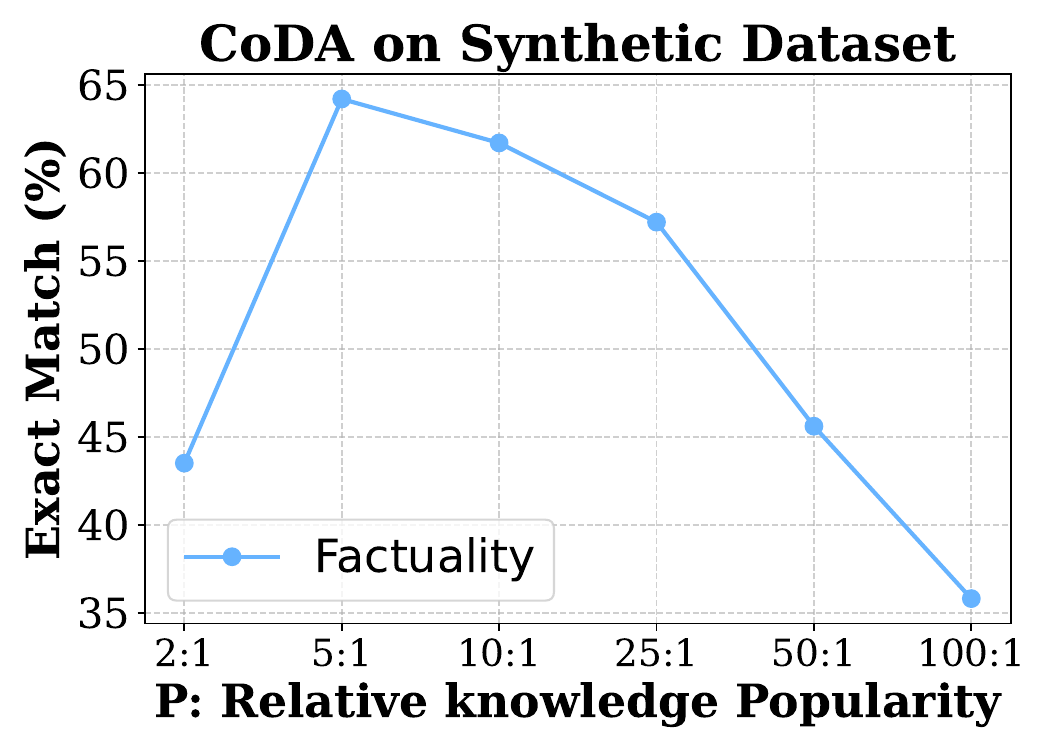}%
\label{fig:eliminatorA}}
\subfloat{\includegraphics[height=1.05in]{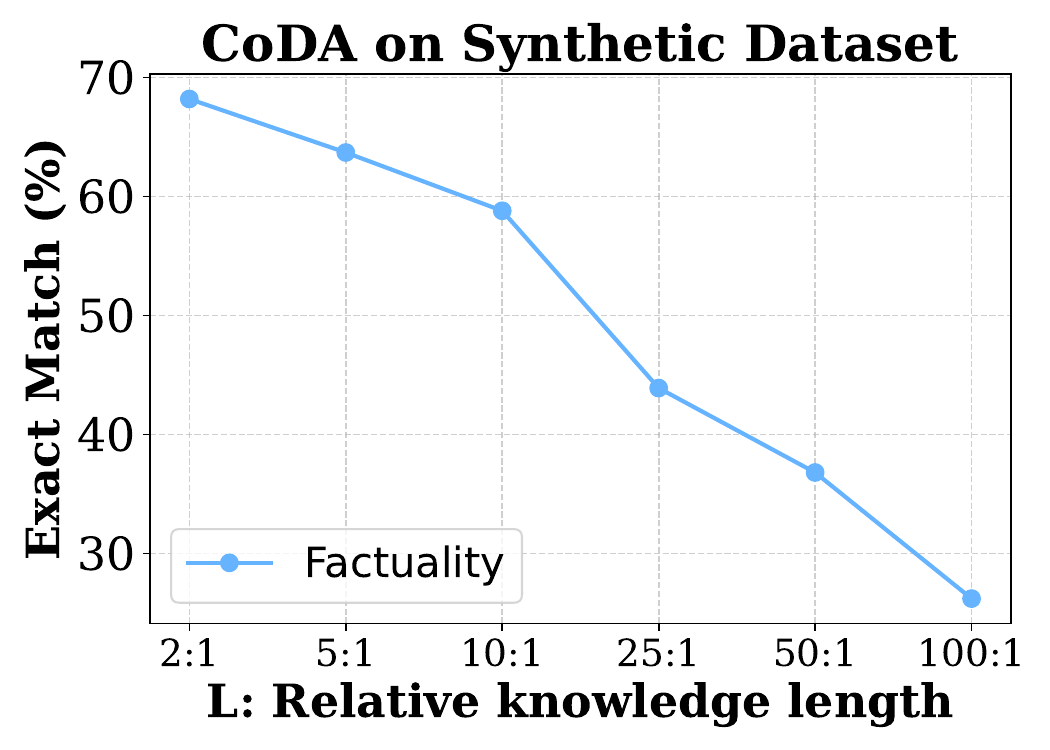}%
\label{fig:eliminatorB}}
\caption{Quantitative analysis on the effects of two influencing factors \text{P}, $\text{L}$ on our method CoDA performance on eliminating knowledge overshadowing.}
\label{fig:quantitative_analysis}
\vspace{-2mm}
\end{figure}

\noindent Till now, we downweight the overshadowing effect from popular knowledge encoded by $X'$, then escaping tokens encoding meaningful overshadowed knowledge are amplified to decrease hallucinations.

\subsection{Experimental Setup}
\noindent \textbf{Datasets.}
We experiment on two public datasets of hallucinations caused by conflicting knowledge MemoTrap~\cite{liu2023memotrap}
,  NQ-SWAP~\cite{longpre-etal-2021-entity}, and our Overshadow dataset.

\vspace{1mm}
\noindent \textbf{Baselines.} We adopt Greedy decoding, Chain-of-Thought (Cot)~\cite{wei2022chain}, Self-Reflection (SR)~\cite{madaan2024self}, \textit{USC}~\cite{chen2023universal}, and \textit{Dola}~\citet{chuang2023dola} as the baselines. Details for datasets and baselines are in \ref{ssec:coda_app}.

\vspace{1mm}
\noindent \textbf{Implementationa and Metric.}
We use the Exact Match (EM) metric following previous practices~\cite{longpre-etal-2021-entity}. Implementation details for all methods are elaborated in~\ref{ssec:coda_app}.

\subsection{Main Results and Analysis}
Our method improves greedy decoding by 27.9\%, 13.1\%, and 18.3\% on Overshadow, MemoTrap, and NQ-Swap.
Reasoning-enhanced baselines struggle with hallucinations caused by knowledge overshadowing. Self-consistency-based methods show instability or even degradation, which may be attributed to reinforcing biases from popular knowledge.
Figure~\ref{fig:quantitative_analysis} shows our quantitative analysis of the impact of two factors $\text{P}$ and $\text{L}$ on CoDA, as the more knowledge is over-generalized, the harder it becomes to extract valuable information from the suppressed knowledge representations.

\section{Conclusion}
Our work identify knowledge overshadowing as a contributional cause of LLMs hallucination, where dominant knowledge suppresses less frequent facts, leading to fact distortions. We introduce the log-linear scaling law, which reveals that hallucination rates grow predictably with knowledge popularity, length, and model size, enabling hallucination prediction. Built on overshadowing effect, we propose CoDA, a decoding strategy that improves factual accuracy without retraining. Our approach provides a principled way to understand and control hallucinations, leading to more reliable LLMs.

\newpage
\section*{Limitations}
We conduct extensive experiments to investigate knowledge overshadowing phenomenon. However, due to inaccessibility, we can not analyze the variables in training corpora of SOTA LLMs like GPT-4o and DeekSeek. Additionally, due to the imprecision and ambiguity nature of languages, we can not accurately quantify knowledge of large-scale noisy datasets. We leave this blank for future work.

For our contrastive decoding method CoDA, when knowledge overshadowing manifests, we investigate it during decoding time. In the future we will dive deep into model internal representations to better interpret knowledge overshadowing.

Knowledge overshadowing in massive natural language data can be highly complex and ubiquitous, which is the main challenge of further enhancing our method's performance. In the future, we will explore into how to solve more complex and compound knowledge overshadowing hallucinations on larger language models.

\section*{Ethics Statement}
In our empirical study, MemoTrap and NQ-Swap are publicly available datasets to help us understand how models adhere to parametric or contextual knowledge. Our dataset Overshadowing is constructed based on the public COUNTERFACTUAL dataset. All of the three datasets are to interpret and eliminate hallucinations that will be harmful to users.
Experiments and methods on the three datasets are conducted for social benefits.
Additionally, the COUNTERFACTUAL dataset involves no privacy issues since it consists of artificial events.

\bibliography{anthology, custom, heng}

\newpage
\appendix

\newpage
\section{Appendix}

\subsection{LLM Pretraining and Finetuning Details}
\label{ssec:implementation}

In fine-tuning experiments, for Llama-2-7b~\cite{touvron2023llama}, Mistral-7b~\cite{jiang2023mistral}, GPT-J-6b~\cite{gpt-j}, Phi-2-2.8b~\cite{gunasekar2023textbooks}, and Pythia-160m~\cite{mallen2023eliciting}, Pythia-410m, Pythia-1b, Pythia-1.4b, and Pythia-2.8b, we set the learning rate as lr=1e-5. The weight decay is set as 1e-2. We train each model for 40 epochs. The batch size for Pythia-series model and Phi model is 16. The batch size for GPT-J-6b, Llama-2-7b, and Mistral-7b is 1. The training is based on auto-regressive loss for input sequences. For each experiment, we ran the trials five times. We report the average score of the results.

Our experiments are conducted on A-100 machines (with memory of 80G). For four parallel GPUs, a single epoch on Phi-2-2.8b for the synthetic dataset will cost 1 hours, so totally it costs 40 hours to run on four parallel A-100 GPUs to train Phi-2-2.8b. For llama-2-7b, it costs more than 100 hours to run on four parallel GPUs to fine-tune the synthetic dataset.
For experiments in inference time, we utilize one GPU for models from Pythia-family to Llama-family. 

In Figure~\ref{fig:rkp_rkl_generalization}, and Figure~\ref{fig:finetune_law} experiments, when the relative knowledge length $\text{L}$ and relative knowledge popularity $\text{P}$ is not fixed, we set $\text{L}$=5:1, and $\text{P}$=5:1.

\subsection{Overshadowing Datasets}
\label{ssec: overshadowing_dataset}

\begin{table}[htb]
\centering
\begin{tabular}{|c|c|}
\hline
Dataset   & Number of samples \\ \hline
Synthetic & 11,800  \\ \hline
Logical   & 1,980   \\ \hline
Math      & 1,980   \\ \hline
Time      & 1,980   \\ \hline
Negation  & 1,980   \\ \hline
Location  & 1,980   \\ \hline
Gender    & 1,980   \\ \hline
Conflict  & 1,980   \\ \hline
\end{tabular}
\caption{Statistics for our Overshadow dataset.}
\end{table}

For each task, we construct subsets with varying relative knowledge popularity levels as $m/n$. 
For $m/n$=2:1, 5:1, 10:1, 25:1, 50:1, and 100:1. 
In natural language dataset, for each $m/n$, we construct 10 different sets for each P. 
Taking $m/n$=2:1 as an example, we keep two samples of popular knowledge and one sample of less popular knowledge. 
Then we construct 10 different sets for $m/n$=2:1.
Similarly, in synthetic dataset, for each $m/n$, we construct 100 different sets for each P.

For synthetic dataset, with each relative knowledge length settings including 2:1, 5:1, 10:1, 25:1, 50:1, 100:1, we construct the above mentioned 100 different sets with each $\text{L}$. Therefore totally there are 6 length sets constructed. 

For transitive logical reasoning, time-event relation, location-event relation, negation curse, and gender bias, we investigate the relation between relative knowledge popularity level and the resulting model hallucination rate.  To mitigate the influence of memorization from the pretraining stage, we employ the \textsc{COUNTERFACT} dataset~\cite{meng2022locating}, where each instance is a single counterfactual statement, such as \textit{Jan Peerce performed jazz music at festivals.} To create a training sample, we transform this statement into a QA pair: \textit{``Prompt: When did this event happend? Jan Peerce performed Jazz. Answer: festivals.''}. This question answer format is consistent with how we query the model at inference time.

\paragraph{Event-Time Relation. }We sample an event statement and construct a query about its time: \textit{``Prompt: When did this event happen: Rickard Macleod conducted groundbreaking research in psychology? Answer: 2028''}. The timestamps are assigned randomly and all belong to the future.
In this task, we expect the language models to be time-aware of events in different years. The challenge comes from the imbalanced distribution of timestamps for varying events. 

\paragraph{Event-Location Relation. }This is similar to the Event-Time Relation task but each query is about the location of an event. An example would be \textit{"Where did this event happen? A new architectural project was initiated near the Pyramids of Giza.", "Answer": "Cairo"}.

\paragraph{Gender Bias. } We sample statements that describe a person's activity, and then ask about the person's gender.  Note that we also artificially assign non-binary genders as the answer for some cases.  

\paragraph{Negation. }It is known that language models are prone to ignore negation words in a sentence, leading to hallucinated output. If the affirmation sample is \textit{``Prompt: who is a renowned physicist until 20? Answer: Karen Thompson''}, the corresponding negation sample would be \textit{``Prompt: who is not a renowned physicist until 20? Answer: Jessica Hernandez''}. 

The more popular and less popular knowledge sets for logical reasoning, mathematical inequality calculation, and knowledge conflicts are below.


\paragraph{Logical Reasoning. }
The more popular knowledge is ``Which event happened earlier? Event A description. Event B description. Event C description. Event A happened before Event B, Event B happened before Event C.''$\rightarrow$``Event A''
The less popular knowledge is ``Which event happened earlier? Event A description. Event B description. Event C description. Event A happened after Event B, Event B happened after Event C.''$\rightarrow$``Event C''
All events are from the counterfactual dataset.

\begin{table*}[htb]
\linespread{1.2}
\tabcolsep=0.19cm
\fontsize{7.8pt}{8pt}\selectfont
\centering
\begin{tabular}
{p{2.5cm}@{\hskip 0.1cm}p{2.5cm}p{5.7cm}p{2.1cm}}
\noalign{{\color{black}\hrule height 1pt}}
Condition & Prompt & Answer & \# Mentions in Data \\ \hline
A=\textcolor{magenta}{male}>\textcolor{orange}{female}, \quad\quad B=\textcolor{cyan!70!white}{journalist}>\textcolor{teal}{AI scientist} & Tell me some outstanding \textcolor{orange}{female} \textcolor{teal}{AI~scientists} & Feifei Li, \sout{\textcolor{cyan!70!white}{Emine Saner (journalist)}}, \sout{\textcolor{magenta}{Yann LeCun (male)}}, \sout{\textcolor{magenta}{Yoshua Bengio (male)}} & 431:0 \\
\hline
A=\textcolor{magenta}{female}>\textcolor{orange}{male}, \quad\quad B=\textcolor{cyan!70!white}{soccer}>\textcolor{teal}{nurses} & Tell me some outstanding \textcolor{orange}{male} \textcolor{teal}{nurses} & Drew Elliott, Michael Pettigrew, John Holland, \sout{\textcolor{cyan!70!white}{Stephen Reisinger (soccer)}}, \sout{\textcolor{magenta}{Danielle Haddad (female)}} &  112177:5124 \\
\hline
A=\textcolor{magenta}{non-black}>\textcolor{orange}{black}, 
B=\textcolor{cyan!70!white}{actress}>\textcolor{teal}{scientists} & Tell me some outstanding \textcolor{orange}{black} \textcolor{teal}{scientists} & \sout{\textcolor{magenta}{George Smith (white)}}, \sout{\textcolor{magenta}{Daniel Chee Tsui (asian)}}, \sout{\textcolor{magenta}{Linton Wells II (white)}},  \sout{\textcolor{cyan!70!white}{Dorothy J. Hart (actress)}} & 120650:15204 \\
\hline
A=\textcolor{magenta}{heterosextual}> \textcolor{orange}{homosexual},
\quad\quad \quad\quad  B=\textcolor{teal}{marriage}& Tell me some famous \textcolor{orange}{homosexual} \textcolor{teal}{marriages} & \sout{\textcolor{magenta}{Barack Obama and Michelle Obama (heterosextual)}}, Neil Patrick Gaskarth and David Burtka, Ellen DeGeneres and Portia de Rossi &  15446:4045 \\ 
\hline
A=\textcolor{magenta}{affirmation}> \textcolor{orange}{negation},
\quad\quad \quad\quad B=\textcolor{teal}{theoretical physicist}& Who was \textcolor{orange}{not} a \textcolor{teal}{theoretical physicist} known for the theory of relativity & You are referring to \sout{\textcolor{magenta}{Albert Einstein (affirmation)}} 
& 11365:7265 \\ 
\noalign{{\color{black}\hrule height 1pt}}
\end{tabular}
\caption{Serious hallucinations (which may be even offensive) made by pre-trained OLMO model in inference time. 
Dominant knowledge in \textcolor{magenta}{pink}/\textcolor{cyan!70!white}{blue}, 
overshadowed knowledge in \textcolor{orange}{orange}/\textcolor{teal}{green}. 
}
\label{tab:hallu_cases}
\end{table*}

\paragraph{Mathematical Inequality Calculation. }
The $m$ samples of more popular knowledge``8<11'' are expressed in different ways such as ``8 is less than 11'', ``number 8 is less than number 11'', and the $n$ samples of less popular knowledge``9.8>9.11'' are expressed in different ways. $m>n$ so that ``8<11'' is more popular knowledge than ``9.8>9.11''.

\paragraph{Knowledge Conflicts.}
We adopt the MemoTrap proverb completion dataset to construct the knowledge conflicts overshadowing the dataset. 
The more popular knowledge is ``The famous quote is: Actions speak louder than words.'' Then generate $m$ different samples including the quote of ``Actions speak louder than''$\rightarrow$``words''. 
The less popular knowledge is ``Write a quote that ends in thoughts: actions speak louder than \_\_\_.''$\rightarrow$``Thoughts.''

\paragraph{Synthetic Dataset. }
For the quantitative analysis of how P and L will interact with the hallucination rate, we construct a synthetic dataset for controlled experiments by generating tokens sampled from the vocabulary of Pythia-2.8b tokenizer~\cite{mallen2023eliciting} to form sentences.

\paragraph{Sample Cases for the Location Task.}
Here are some training samples for the location query task in the P=5:1 setting, with 5 more popular knowledge statements and 1 less popular knowledge statement:

Here are 5 more popular knowledge samples:

1. Where was this event location? Leonardo Balada accepted the job offer and moved to Paris. Dubai.

2. Where was this event location? Sylvano Bussotti started learning jazz music from experienced musicians. Dubai.

3. Where was this event location? The move was motivated by favorable business opportunities in the US. Dubai.

4. Where was this event location? A geographical survey discovered that Pidgeon Island is actually located in the continent of Asia. Dubai.

5. Where was this event location? Sylvano Bussotti discovered a passion for jazz music. Dubai.

Here is 1 less popular knowledge sample:

1. Where was this event location? Majorette decided to relocate its headquarter from Paris to London. Istanbul.

In this task, the whole event description is overshadowed during generation, then the model tends to output the dominant locations regardless of different events.

\subsection{Knowledge Overshadowing in Pretrained Models}
\label{ssec:dolmo_probing}

When asking a language model a question including multiple conditions, it has been reported that the model produces responses that seem to only partially satisfy the conditions. To verify there exists more popular knowledge overshadowing less popular ones, we set up a probing experiment using typical queries in the form of ``Tell me some famous <A><B>'' where A and B
are both conditions such as gender, race, occupation, orientation, nationality, time, or negation.
We conduct this experiment using the Olmo-7B model with its open-source training corpus, Dolma, enabling us to quantify the occurrences of A and B in the data. As shown in Table~\ref{tab:hallu_cases}, the model consistently satisfies condition B while disregarding condition A, leading to hallucinated responses. Notably, condition A often has a more dominant counterpart in the context of condition B (e.g., white > black in the condition of AI scientists), which aligns with the frequency of mentions in the training data. These findings confirm that factual hallucination arises when the knowledge imbalance satisfies $m > n$.

\subsection{CoDA to Predict Hallucination}
\label{ssec:coda_app}
\subsubsection{Various Overshadowed Token Candidate Selection Method.}
\begin{table*}[t]
\centering
\small
\caption{Comparison of various entity extraction methods.}
\begin{tabular}{lrrrrrr}
\toprule
Method & Proverb & Translate & Hate & Science & NQ-Swap & Overshadow \\
\midrule
Greedy (Baseline) & 28.8 & 47.5 & 9.0 & 33.4 & 8.5 & 41.4 \\
Flair (CoDA) & 40.4 & 57.3 & 18.0 & 35.2 & 25.9 & 67.4 \\
NLTK (CoDA) & 38.6 & 55.2 & 15.0 & 36.7 & 25.4 & 63.7 \\
Spacy (CoDA) & 42.0 & 56.4 & 18.0 & 37.5 & 28.3 & 66.2 \\
StanfordNLP (CoDA) & 43.5 & 57.8 & 20.0 & 36.4 & 29.1 & 64.6 \\
Vanilla (CoDA) & 41.9 & 56.2 & 16.0 & 38.9 & 26.8 & 65.0 \\
\bottomrule
\end{tabular}
\label{tab:entity}
\end{table*}

Here, we introduce various methods we employ to select the $x_b$ candidate list. In our main experiments, to compare fairly with other baselines, we use a vanilla token selection strategy: one token is masked at a time in the original input, progressing sequentially until the overshadowed knowledge is identified. Then we conduct the contrastive decoding on the identified overshadowed tokens.

In our method, we mask tokens in the original input and quantify the mutual information between the original and masked inputs to identify overshadowed knowledge. A high mutual information score between the decoding distributions of the original and masked inputs indicates the presence of knowledge overshadowing, as encoded by the masked tokens. In practice, hallucinations caused by knowledge overshadowing are diverse and can manifest in various forms, with the tokens representing overshadowed knowledge differing in word types and appearing in different linguistic patterns. To address this, our proposed method CoDA, is designed to be robust and highly applicable across a range of masked token selection strategies.
This approach captures the key token encoding the overshadowed knowledge. Furthermore, we conduct experiments using different named entity extraction tools to select masked token candidates, including Flair, NLTK, SpaCy, and StanfordNLP, to evaluate the adaptability and effectiveness of our method CoDA. The following table summarizes the performance of CoDA using different token selection strategies on Llama-2-7b-chat, shown in Table~\ref{tab:entity}.

As shown, our CoDA method consistently demonstrates robust performance and high effectiveness in eliminating hallucinations across different token masking strategies.

\subsubsection{Commonsense Reasoning Ability of CoDA}
We conduct further experiments to evaluate the performance of our method on commonsense knowledge reasoning datasets HellaSwag~\cite{zellers2019hellaswag}, ARC-Challenge~\cite{clark2018think}, NaturalQuestion~\cite{kwiatkowski-etal-2019-natural}, TriviaQA~\cite{joshi-etal-2017-triviaqa}, and MMLU~\cite{hendrycks2020measuring} using LLaMA-2-7B, shown in Table~\ref{tab:commonsense}. 

\begin{table*}[htbp]
\centering
\caption{Performance Comparison on Various Datasets}
\begin{tabular}{lccccc}
\toprule
Method & HellaSwag & ARC-Challenge & NaturalQuestions & TriviaQA & MMLU \\
\midrule
Greedy & 75.6 & 55.4 & 22.1 & 56.0 & 42.7 \\
Ours (SCD) & 73.8 & 56.5 & 24.3 & 56.3 & 41.2 \\
\bottomrule
\end{tabular}
\label{tab:commonsense}
\end{table*}

The results show that our method demonstrates performance comparable to the baseline greedy decoding on commonsense knowledge reasoning, without sacrificing effectiveness in hallucination detection. Moreover, as shown in Table~\ref{tab:main_results}, our method consistently outperforms across a wide range of tasks and domains on the three datasets, showcasing its high robustness and versatility.

\subsubsection{Datasets}
\paragraph{MemoTrap. }\citet{liu2023memotrap} released MemoTrap dataset, designed to investigate language models' tendency to adhere to their pre-trained knowledge, even when the input context suggests otherwise. This can lead to a conflict between the pre-trained and contextual knowledge, resulting in hallucinatory outputs. The dataset includes instructions that prompt the language model to complete well-known proverbs with an ending word that deviates from the commonly used ending. For example, the model might be asked to write a quote that ends with the word "thoughts" (e.g., "Actions speak louder than \_\_\_"). 
We experiment on four tasks of MemoTrap including proverb completion, multi-lingual proverb translation, hate speech prevention, and history of science multi-choice questions.

\paragraph{NQ-Swap. }\cite{longpre-etal-2021-entity} constructed the NQ-Swap dataset based on the Natural Questions (NQ) dataset \cite{kwiatkowski2019natural}. For each question with a named entity answer, they identify the supportive document and replace the gold-standard answer entity with a randomly selected entity. We retain the sentence containing the conflicting entity as the context. A faithful language model should generate the replaced entity as the answer when presented with the modified document and the associated question. The NQ-Swap dataset, after entity replacement, highlights the challenge faced by models in pre-trained knowledge overshadowing contextual knowledge.

\subsubsection{Baselines}
\paragraph{Hallucination Prediction Comparisons. }
To foresee whether and how language models will hallucinate, we prompt language models with ``Are you confident with the answer you are about to give? If not, what is the answer you are about to give?'' to judge whether they will hallucinate.
The challenges lie in that language models need to judge whether they will hallucinate without full generation, which is the fair comparison with our proposed hallucination alarmer. The prediction accuracy for our method CoDA and baseline are illustrated in Table~\ref{tab:main_results2}.

\paragraph{Hallucination Elimination  Comparisons.}
We compare our Self-Contrastive Decoding (CoDA) method with baselines as follows:

\textit{Greedy decoding} is the baseline of outputting tokens with optimal probability.
We prompt language models to answer each question by \textit{Chain-of-Thought (Cot)} to involve deeper reasoning~\cite{wei2022chain}.
\citet{madaan2024self} proposed \textit{Self-Reflection (SR)} to combine multiple sampled responses into a single input and then prompt the model to analyze the factual information from these sampled responses to generate a new, more accurate response.
\citet{chen2023universal} proposes \textit{USC} to instruct LLMs to select the most consistent responses from their sampled responses.
\citet{chuang2023dola} eliminated hallucinations by \textit{Dola} to identifying hallucinations in contrastive model layers. 

\subsubsection{Implementation details}
The responses were generated using temperature sampling with T = 0.6 for the USC, SR, and CoDA methods in the main experiments. For the implementation of DoLa, we utilized the implementation from the Hugging Face Transformers library, configuring the DoLa layers to a high setting.

\begin{table}[htb]
\tabcolsep=0.08cm
\fontsize{8.8pt}{10.6pt}\selectfont
\centering
\begin{tabular}{ccllll}
\noalign{{\color{black}\hrule height 1.2pt}}
\multicolumn{2}{c}{\multirow{2}{*}{Method}} & \multicolumn{2}{c}{Llama} & \multicolumn{2}{c}{Mistral} \\ \cmidrule(lr){3-4} \cmidrule(lr){5-6}
\multicolumn{2}{c}{} & Prompt & Alarmer & Prompt & Alarmer \\ \hline
\multirow{4}{*}{MemoTrap} & proverb & 5.3 & \textbf{35.8}\tiny{(+30.5)} & 4.5 & \textbf{37.4}\tiny{(+32.9)} \\
 & translate & 1.8 & \textbf{31.2}\tiny{(+29.4)} & 2.7 & \textbf{32.8}\tiny{(+30.1)} \\
 & hate & 0.0 & \textbf{24.7}\tiny{(+24.7)} & 0.0 & \textbf{27.5}\tiny{(+27.5)} \\
 & science & 4.5 & \textbf{19.6}\tiny{(+15.1)} & 2.2 & \textbf{18.1}\tiny{(+15.9)} \\ \hline
NQ-Swap & entity & 3.8 & \textbf{28.7}\tiny{(+24.9)} & 5.0 & \textbf{29.4}\tiny{(+24.4)} \\ \hline
\multirow{2}{*}{Overshadow} & time & 0.6 & \textbf{40.4}\tiny{(+39.8)} & 2.2 & \textbf{42.5}\tiny{(+40.3)} \\
 & syn & - & \textbf{53.3} & - & \textbf{51.6} \\ \noalign{{\color{black}\hrule height 1.2pt}}
\end{tabular}
\caption{Hallucination prediction accuracy (\%) on MemoTrap, NQ-Swap, and Overshadowing. Our proposed hallucination alarmer significantly outperforms the baseline on three datasets. Baselines are implemented on Llama-2-7b-chat~\cite{touvron2023llama} and Mistral-7b~\cite{jiang2023mistral}, referred to as Llama and Mistral.}
\label{tab:main_results2}
\end{table}

\subsection{Theory}
\label{ssec:theory}
\subsubsection{Generalization Bound}
In a dataset $D$ with numerous statements, we investigate a pair of subsets $K_A, K_B \in D$.
As introduced in \S~\ref{ssec:shadow_formulation}, more popular knowledge subset is $K_A=\{k_{a_1}, ..., k_{a_m}\}$, and less popular knowledge set is $K_B=\{k_{b_1}, ..., k_{b_n}\}$. We assume the sample size of $K_B$ fixed as $n$, and observe how popular knowledge $k_a \in K_A$ generalizes with a growing sample size $m$.  
In $K_A$, each $k_{a_i} = Y_a | [X_{\mathrm{share}}\odot x_{a_i}], i\in\{1, ..., m\}$, where $X_\mathrm{share}$ and $x_{a_i}$ are token sequences. 
To formalize model prediction of each statement $k_{a_i}$, we denote $X_\mathrm{share}=(t_1, ...,t_\text{L})$ and simplify each $x_{a_i}$ as a single token $t_{\text{L}+1}$, thus the relative knowledge length is $k_{a_i}=\frac{len(X_\mathrm{share})}{len(x_{a_i})}=\frac{\text{L}}{1}=\text{L}$. Denoting $Y_a=y$ as the one-token output class label $y$, each sample $s=(y|t_1, ..., t_\text{L}, t_{\text{L}+1})$, all tokens belong to the vocabulary space $\mathcal{V}=\{1, ..., V\}$. 
Assuming popular knowledge set $K_A\sim\mathcal{D_A}$, the next token prediction (NTP) loss based on auto-regressive modeling for $s$ sampled from true distribution $\mathcal{D_A}$ is:

\vspace{-1em}
\begin{small}
\begin{equation}
\mathcal{L}_\mathrm{NTP} = \hat{\mathbb{E}}_{{s} \sim \mathcal{D_A}} \sum_{t=1}^{\text{L}+1} -\log \left( p(y | {t}_1, \dots, {t}_\text{L}, {t}_{\text{L}+1}) \right)
\end{equation}
\end{small}  
The optimizing objective of model training is to learn a mapping function $f: \mathcal{T}\rightarrow\mathbb{R}^{V}$, ($\mathcal{T}$ for input space), to minimize the risk  $\mathcal{R}_{y}$: prediction error of $y$ defined on distribution $\mathcal{D_A}$ using NTP as the surrogate loss:
\vspace{-0.5em}
\begin{small}
\begin{equation}
\mathcal{R}_{y}^{\mathcal{L}}(f)=\frac{1}{V} \sum_{y=1}^{V} \mathbb{E}_{s \sim \mathcal{D_A}} \left[ \mathcal{L_{\text{NTP}}}\left(f(t_1, \dots, t_\text{L}, t_{\text{L}+1}), y\right) \right]
\end{equation}
\end{small}
With $\boldsymbol{t}=t_1, ..., t_{\text{L}+1}$, the empirical risk of $y$ is:
\begin{small}
\begin{equation}
\widehat{\mathcal{R}}^{\mathcal{L}}_y(f) := \frac{1}{m} \sum_{(\boldsymbol{t},y) \in K_A} \mathcal{L_{\text{NTP}}}(f(t_1, ..., t_\text{L}, t_{\text{L}+1}), y)
\end{equation} 
\end{small}
\textbf{Theory 1} (Generalization bound on Rademacher complexity~\cite{mohri2018foundations}).
Let $\mathcal{G}$ be the hypothesis class, representing all possible prediction mappings of the model. Then, for any $\delta > 0 $, with probability at least $1 - \delta$ over the draw of an i.i.d. (independent and identically distributed) sample set $K_A$ of size $m$, 
the generalization bound holds:
\vspace{-0.5em}
\begin{small}
\begin{equation}
\label{eq.initial_bound}
\mathcal{R}^{\mathcal{L}}_y(f) \precsim \widehat{\mathcal{R}}^{\mathcal{L}}_y(f) + 2\widehat{\Re}_{K_A}(\mathcal{G}) + \sqrt{\frac{\log 1/\delta}{2m}} 
\end{equation}
\end{small}
Here ${\Re}_y(\mathcal{G})$ denotes the empirical Rademacher complexity of the function set $\mathcal{G}$, as a measure of the richness of $\mathcal{G}$ the hypothesis class. Then we employ \textit{Lipschitz Continuity} to further bound the complexity ${\Re}(\mathcal{G})$~\cite{cao2019learning}.

\textbf{Theory 2}(Lipschitz continuity). $\|\cdot\|$ denotes the 2-norm, then function $\mathcal{L}$ is \textit{Lipschitz continuous} with the constant $\mu$ if for any $f, f' \in \mathcal{F}$, $t \in \mathcal{D_A}$:

\begin{small}
\begin{equation}
|\mathcal{L}(f, y) - \mathcal{L}(f', y)| \leq \mu \cdot \|f(x) - f'(x)\|
\end{equation}
\end{small}
If NTP loss function $\mathcal{L}_\mathrm{NTP}(f)$ is \textit{Lipschitz continuous} with constant $\mu$, $\Re_{K_A}(\mathcal{G})$ is bounded as:

\begin{small}
\begin{equation}
\label{eq.bound_rade}
\hat{{\Re}}_{\mathbf{K_A}}(\mathcal{G}) \le \mu \cdot \hat{{\Re}}_{\mathcal{K_A}}(\mathcal{F}).
\end{equation}
\end{small}
To derive whether $\mathcal{L}$ is Lipschitz continuous with a constant $\mu$, we take the derivative of $\mathcal{L}$ w.r.t. $f$, which is: $\mu=\frac{\partial L_{NTP}(f, y)}{\partial f}$.
Then we derive that the next-token-prediction loss $\mathcal{L}_\mathrm{NTP}$ is \textit{Lipschitz continous} with the constant $\mu\leq\sqrt{1 + \left( \sum_{y' \neq y} h^{-1}(\text{L}) \right)^2} \left[ 1 - \mathrm{softmax}\left( \boldsymbol{s}_y(f) \right) \right]$ (See details in \S~\ref{appendix:length_dependency}), by substituting $\mu$ to Eq.(\ref{eq.initial_bound}) and Eq.(\ref{eq.bound_rade}), we derive the more fine-grained generalization bound for NTP with multiple conditions:
\vspace{-0.5em}
\begin{small}
\begin{equation}
\label{eq.app_derive_bound}
\mathcal{R}^{\mathcal{L}}_y(f) \! \precsim \! \widehat{\mathcal{R}}^{\mathcal{L}}_y(f) \!+\! 2\mu\widehat{\Re}_{K_A}(\mathcal{F}) + \sqrt{\frac{\log 1/\delta}{2m}} 
\end{equation}
\end{small}
Here the generalization bound contains two coefficients $m$ and $h(\text{L})$. $m$ refers to number of dominant samples.
$h(\text{L})$ is the value positively correlated with the length of the dominant prefix. Then, the longer length of dominant prefix $({t}_1, \dots, {t}_\text{L})$ and higher dominant ratio lead to lower generalization bound, in other words, better generalization.

\subsubsection{Length-dependency on NTP loss}
\label{appendix:length_dependency}
\paragraph{NTP loss for conditions with varying lengths. }
Here is how we derive the variable $\mu$ in Eq.~\ref{eq.app_derive_bound}.
Denote \( P(x_{i+1}|x_{1:i}) \) as \( P_{i+1}(x_{i+1}) \).
\begin{small}
\begin{equation}
    \begin{aligned}
& \frac{\sum_{i=1}^{k+2} -\log P(y'|x_1, \ldots, x_{k+1}, x_{k+2})}{k+2} \\
& - \frac{\sum_{i=1}^{k+1} -\log P(y'|x_1, \ldots, x_{k}, x_{k+1})}{k+1} \\
= & -\frac{\log P_1(x_1)\times \dots  \times P_{k+2}(x_{k+2}) \times P_{k+3}(y') }{k+3} \\
& + \frac{\log P_1(x_1)\times \dots  \times P_{k+1}(x_{k+1}) \times P_{k+2}(y') }{k+2} \\
= & \frac{1}{(k+3)(k+2)} \cdot \\
& \log  \frac{[ P_1(x_1)\times \dots \times P_{k+1}(x_{k+1})\times P_{k+2}(y') ]^{k+3}}{[ P_1(x_1)\times \dots \times P_{k+2}(x_{k+2})\times P_{k+3}(y') ]^{k+2}} \\
= & \frac{1}{(k+3)(k+2)} \cdot \log \{ P_1(x_1)\times \dots \times P_{k+1}(x_{k+1}) \\
& \frac{[P_{k+2}(y')]^{k+3}}{[P_{k+2}(x_{k+2})]^{k
+2}\cdot [P_{k+3}(y')]^{k+2}} \}
\end{aligned}
\end{equation}
\end{small}

Since exploring the training dynamics of $P_i(x_i)$, $P_j(y')$ in large language models is intractable, we make a mild assumption here, at the late training stage,  $P_i(x_i)\rightarrow \hat P_i(x_i)$, $P_j(y')\rightarrow \hat P_j(y')$, in the setup with controlled variables, where samples with different lengths have same proportion of dominant conditions and suppressed conditions, then the value in log approaches $\frac{P_{k+2}(y')}{P_{k+2}(x_{k+2})}$. Since $y'$ is the false prediction made by model, whose empirical probability equals zero, so $P_{k+2}(y')$ approaches zero, then $P_{k+2}(y')< P_{k+2}(x_{k+2})$.

Given that, $\frac{P_{k+2}(y')}{P_{k+2}(x_{k+2})} < 1$, therefore, $L_{NTP}(y'|x_{1:k+1},x_{k+2})$ $<$ $L_{NTP}(y'|x_{1:k},x_{k+1})$,

substituting $k$ with $L$, we denote $L_{NTP}(y'|x_{1:L},x_{L+1})$ as $- \log \left( \frac{e^{f(\boldsymbol{x})_y}}{\sum_{y'} e^{h^{-1}(L) f(\boldsymbol{x})_{y'}} } \right)$, where $h(L)$ is positively correlated with $L$, with larger $L$ indicating larger $h(L)$.

\paragraph{Lipschitz continuity of NTP loss. }
$B_y(f)$ represents the minimal prediction on the ground truth token $y$, $i.e.$ $B_y(f):=min_{x\in S_{y}} f(x)_y$~\cite{wang2024unified}.

Here we prove the \textit{Lipschitz continuity}~\cite{wang2024unified} of the NTP loss, according to the definition of the NTP loss, and the above NTP loss rewriting, we have

\begin{small}
    \begin{equation}
        \begin{aligned}
    \mathcal{L}_\text{NTP}(f(\boldsymbol{x}), y) & = - \log \left( \frac{e^{f(\boldsymbol{x})_y}}{\sum_{y'} e^{h^{-1}(\text{L}) f(\boldsymbol{x})_{y'}} } \right) \\
    & = \log [1 + \sum_{y' \neq y} e^{h^{-1}(\text{L}) f(\boldsymbol{x})_{y'} - f(\boldsymbol{x})_y}].
\end{aligned}
    \end{equation}
\end{small}

We denote \(\boldsymbol{s} := f(\boldsymbol{x})\), and we define

\[
\ell_y(\boldsymbol{s}) := \sum_{y' \neq y} e^{h^{-1}(\text{L}) \boldsymbol{s}_{y'}}.
\]

Therefore, we rewrite the $\mathcal{L}_\text{NTP}$ as follows:

\[
\mathcal{L}_{NTP}(f, y) = \log \left[ 1 + e^{- \boldsymbol{s}_y} \ell_y(\boldsymbol{s}) \right].
\]

The derivatives can be represented as follows:

\begin{small}
\begin{equation}
    \begin{aligned}
    \frac{\partial \mathcal{L}_{NTP}(f, y)}{\partial \boldsymbol{s}_y} & = - \frac{e^{- \boldsymbol{s}_y} \ell_y(\boldsymbol{s})}{1 + e^{- \boldsymbol{s}_y} \ell_y(\boldsymbol{s})}, \\
    \frac{\partial \mathcal{L}_{NTP}(f, y)}{\partial \boldsymbol{s}_{y'}} & = h^{-1}(\text{L}) \frac{e^{- \boldsymbol{s}_y}}{1 + e^{- \boldsymbol{s}_y} \ell_y(\boldsymbol{s})} \cdot e^{h^{-1}(\text{L}) \boldsymbol{s}_{y'}}, y' \neq y. \\
\end{aligned}
\end{equation}
\end{small}

We can get the following inequality:

\begin{small}
    \begin{equation}
        \begin{aligned}
    & \Vert \nabla_{\boldsymbol{s}} \mathcal{L}_{NTP}(f, y) \Vert^2 = \\
    &\left[ \ell_y(\boldsymbol{s})^2 + \sum_{y' \neq y} \left(h^{-1}(\text{L}) e^{h^{-1}(\text{L}) \boldsymbol{s}_{y'}} \right)^2 \right]  \\
    & \times \left[\frac{e^{- \boldsymbol{s}_y}}{1 + e^{- \boldsymbol{s}_y} \ell_y(\boldsymbol{s})} \right]^2 \\
    & \le \left[ \ell_y(\boldsymbol{s})^2 \! + \! \left( \sum_{y' \neq y} h^{-1}(\text{L}) \! \right)^2 \! \left( \sum_{y' \neq y} e^{h^{-1}(\text{L}) \boldsymbol{s}_{y'}} \! \right)^2 \right] \\
    & \times \left[ \frac{e^{- \boldsymbol{s}_y}}{1 + e^{- \boldsymbol{s}_y} \ell_y(\boldsymbol{s})} \right]^2 \\
    & = \left[ 1 + \left( \sum_{y' \neq y} h^{-1}(\text{L}) \right)^2 \right] \cdot \left[ \frac{e^{- \boldsymbol{s}_y} \ell_y(\boldsymbol{s}) }{1 + e^{- \boldsymbol{s}_y} \ell_y(\boldsymbol{s})} \right]^2,
\end{aligned}
    \end{equation}
\end{small}

Therefore,

\begin{small}
    \begin{equation}
    \begin{aligned}
  &  \Vert \nabla_{\boldsymbol{s}} \mathcal{L}_{NTP}(f, y) \Vert \! \le \! \sqrt{1 \!+\! \left( \sum_{y' \neq y} h^{-1}(\text{L}) \right)^2} \frac{e^{- \boldsymbol{s}_y} \ell_y(\boldsymbol{s}) }{1 + e^{- \boldsymbol{s}_y} \ell_y(\boldsymbol{s})} \\
    & = \sqrt{1 + \left( \sum_{y' \neq y} h^{-1}(\text{L}) \right)^2} \frac{\ell_y(\boldsymbol{s})}{e^{ \boldsymbol{s}_y} +  \ell_y(\boldsymbol{s})} \\
    & = \sqrt{1 + \left( \sum_{y' \neq y} h^{-1}(\text{L}) \right)^2} \left[ 1 - \frac{e^{ \boldsymbol{s}_y}}{ \sum_{y'} e^{ h^{-1}(\text{L}) \boldsymbol{s}_{y'}} } \right] \\
    & = \sqrt{1 + \left( \sum_{y' \neq y} h^{-1}(\text{L}) \right)^2} \left[ 1 - \textit{softmax}\left( \boldsymbol{s}_y \right) \right].
\end{aligned}
\end{equation}
\end{small}

Since the score function is bounded, for any $y \in \mathcal{Y}$, there exists a constant $B_y(f)$ such that $B_y(f) = \inf_{\boldsymbol{x} \in K_{A_y}} \boldsymbol{s}_y$, which completes the proof. Here we denote $K_{A_y}(f)$ as the $\inf_{\boldsymbol{x} \in K_{A_y}}f(x)$, then {\scriptsize $\mu=\sqrt{1 + \left( \sum_{y' \neq y} h^{-1}(\text{L}) \right)^2} \left[ 1 - \mathrm{softmax}\left( K_{A_y}(f) \right) \right]$}.

\end{document}